\documentclass[review]{elsarticle}

\usepackage{lineno,hyperref}

\usepackage{times}
\usepackage{soul}
\usepackage{url}

\usepackage[utf8]{inputenc}
\usepackage[small]{caption}
\usepackage{graphicx}
\usepackage{amsmath,amssymb,amsthm}
\usepackage{booktabs}
\usepackage{algorithm}
\usepackage{algorithmic}
\usepackage{multirow}
\usepackage{subfigure}
\usepackage{color}
\newtheorem{theorem}{Theorem}

\newtheorem{remark}{Remark}

\journal{Journal of \LaTeX\ Templates}

\begin{document}

\begin{frontmatter}

\title{Expand Globally, Shrink Locally: Discriminant Multi-label Learning with Missing Labels}
\tnotetext[mytitlenote]{Fully documented templates are available in the elsarticle package on \href{http://www.ctan.org/tex-archive/macros/latex/contrib/elsarticle}{CTAN}.}


\author[mymainaddress,mysecondaryaddress]{Zhongchen Ma}

\author[mysecondaryaddress,mythirdaddress]{Songcan Chen\corref{mycorrespondingauthor}}
\cortext[mycorrespondingauthor]{Corresponding author}
\ead{s.chen@nuaa.edu.cn}

\address[mymainaddress]{The School of Computer Science \& communications Engineering, Jiangsu University, Zhenjiang 212013, China}
\address[mysecondaryaddress]{College of Computer Science and Technology, Nanjing University of Aeronautics and Astronautics (NUAA), Nanjing 211106, China}
\address[mythirdaddress]{MIIT Key Laboratory of Pattern Analysis and Machine Intelligence}

\begin{abstract}

In multi-label learning, the issue of missing labels brings a major challenge. Many methods attempt to recovery missing labels by exploiting low-rank structure of label matrix. However, these methods just utilize global low-rank label structure, ignore both local low-rank label structures and label discriminant information to some extent, leaving room for further performance improvement. In this paper, we develop a simple yet effective discriminant multi-label learning (DM2L) method for multi-label learning with missing labels. Specifically, we impose the low-rank structures on all the predictions of instances from the same labels (local shrinking of rank), and a maximally separated structure (high-rank structure) on the predictions of instances from different labels (global expanding of rank). In this way, these imposed low-rank structures can help modeling both local and global low-rank label structures, while the imposed high-rank structure can help providing more underlying discriminability. Our subsequent theoretical analysis also supports these intuitions. In addition, we provide a nonlinear extension via using kernel trick to enhance DM2L and establish a concave-convex objective to learn these models. Compared to the other methods, our method involves the fewest assumptions and only one hyper-parameter. Even so, extensive experiments show that our method still outperforms the state-of-the-art methods.

\end{abstract}


\end{frontmatter}


\section{Introduction}\label{sec:introduction}

Multi-label learning, handling instance associated with multiple labels, has attracted lots of attention due to its widespread applicability in diverse fields such as image annotations \cite{boutell2004learning}, music classification \cite{turnbull2008semantic}, multi-topic text categorization \cite{ueda2003parametric}, etc. During the past decades, a large number of methods have been proposed and achieved good performance for multi-label learning. According to \cite{tsoumakas2012introduction}, these methods can be roughly divided into two categories: algorithm adaptation and problem transformation. Algorithm adaption methods attempt to adapt popular learning techniques  to handle multi-label learning problems directly. Some notable examples include ML-$k$NN \cite{zhang2007ml}, ML-DT \cite{clare2001knowledge} and Rank-SVM \cite{elisseeff2002kernel}. While problem transformation methods tackle the problem by transforming it to other well-established learning scenarios. Binary Relevance \cite{boutell2004learning}, Classifier Chains \cite{read2011classifier}, Calibrated Label Ranking \cite{furnkranz2008multilabel} and Random  $k$-labelsets \cite{tsoumakas2007random} fall into this category.

The aforementioned methods generally assume that the labels of training instances are complete. Unfortunately, in real-world applications, some labels tend to miss from the training set, consequently forming a kind of weakly supervised learning problem \cite{zhou2017brief}. Label missing can generally be due to that human labelers may sometimes ignore labels they do not know or of little interest, or following the guide by some algorithms to reduce labeling costs \cite{huang2015multi,gao2016multi}. Therefore, these methods will fail in this situation.


To solve this problem, a simple solution is to discard all samples with missing labels, though at the expense of potentially losing a significant amount of label information. Another is to recover the missing labels by exploiting label structure, and of which the low-rank of label matrix is the most commonly used label structure due to theoretical support, namely, for a low-rank matrix $\mathbf{M}\in \mathbb{R}^{n \times m}$ of rank $r$, it can be perfectly recovered from $O\left(r(n+m) \ln ^{2}(n+m)\right)$ observed entries when the observed entries are uniformly sampled from the $\mathbf{M}$. Besides, when additional label structures are incorporated into learning of low-rank models, they can achieve better or even the state-of-the-art classification performance. For example, \cite{yang2016improving} incorporates structured semantic correlations into low-rank model learning and gets improved performance. 

\begin{figure}[htp!]
	\setlength{\abovecaptionskip}{0.2cm}
	\setlength{\belowcaptionskip}{0.cm}
	\centering
	\includegraphics[width=0.6\textwidth]{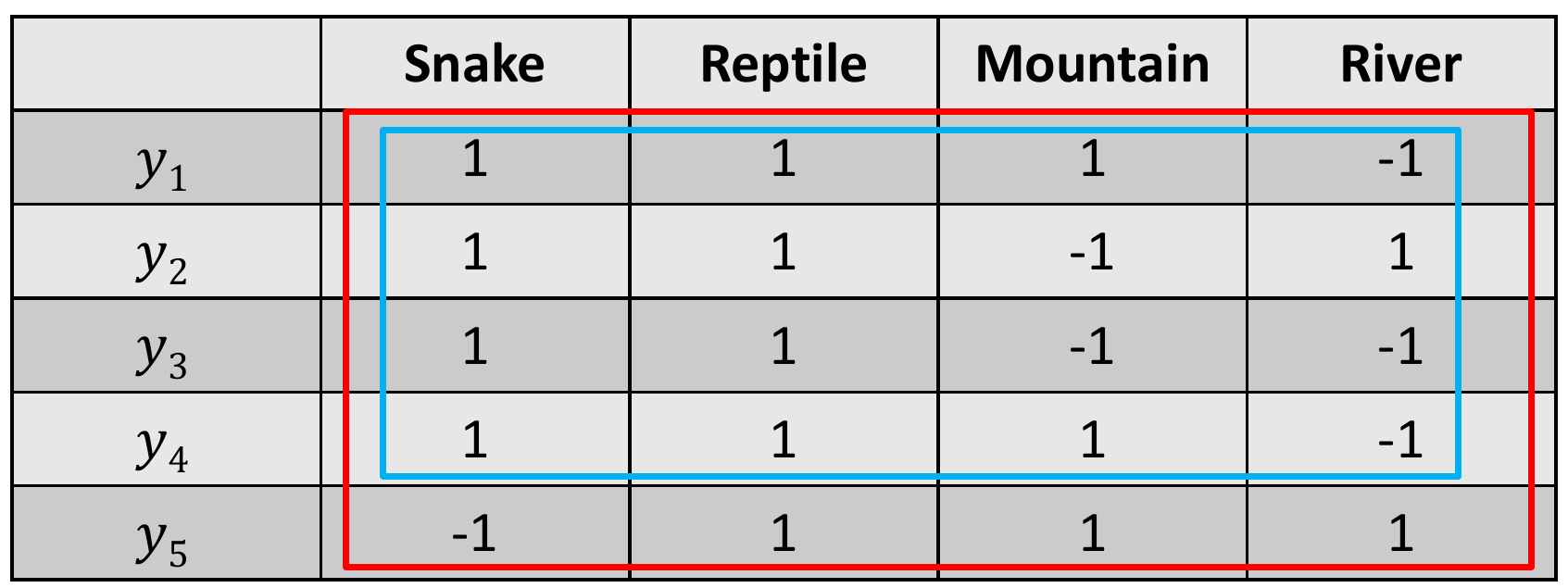}
	\caption{An example of local low-rank label structure. The label matrix in the figure is taken from a real multi-label dataset "Corel16k".  As we can see, if a sample is labelled as "snake", then this sample must be labelled as "reptile", but the opposite is not true. As a result, although the rank of the original label matrix (inside the red box) is $4$, the rank of the label submatrix labelled as "snake" (inside the blue box) is $3$.}
	\label{local_lrls}
\end{figure}

In multi-label learning, \emph{asymmetric} co-occurrence relation is one of the most useful label structures. To be more specific, if a sample is labelled $\lambda_1$, then it must be labelled $\lambda_2$, but the opposite is not necessarily true. If all the samples labelled $\lambda_1$ are used to form a new sample submatrix, then the corresponding label submatrix has smaller rank than the original label matrix. An example of this phenomenon is shown in Fig. \ref{local_lrls}. Based on this, we call \emph{asymmetric} co-occurrence label structure as local low-rank label structure, while the low-rank property of whole label matrix as global low-rank label structure.

\begin{figure}[htp!]
	\setlength{\abovecaptionskip}{0.2cm}
	\setlength{\belowcaptionskip}{0.cm}
	\centering
	\includegraphics[width=0.6\textwidth]{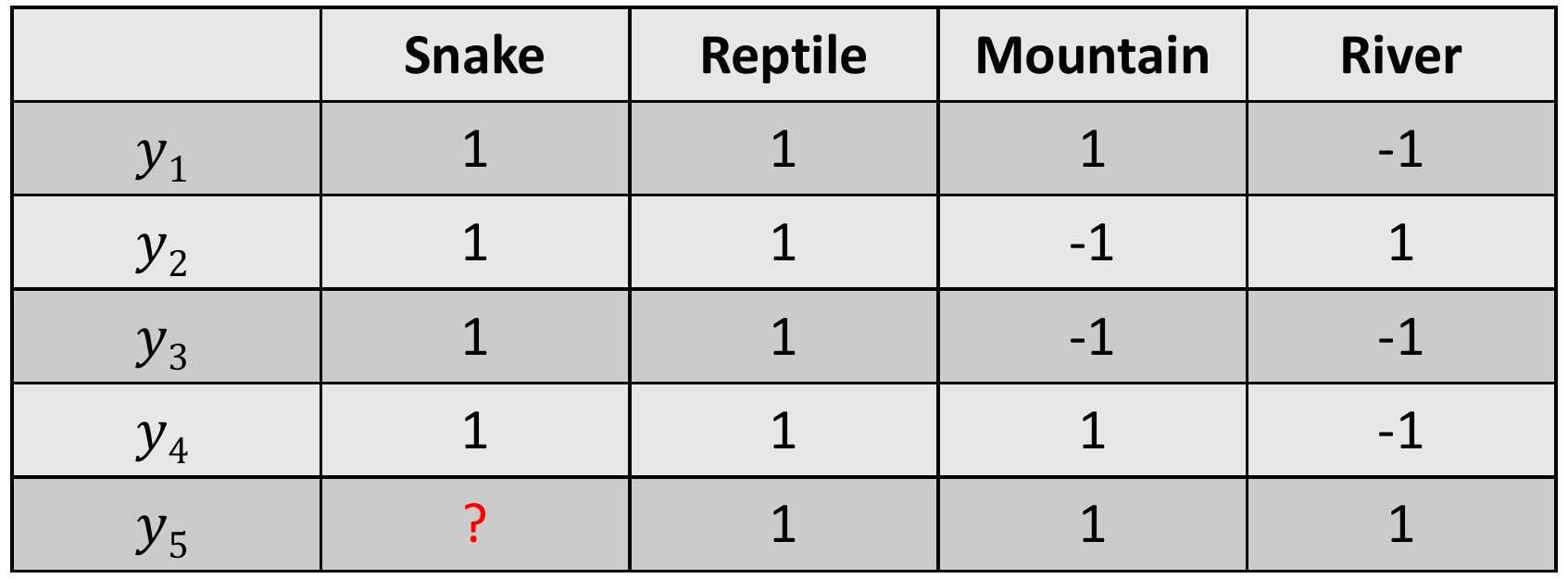}
	\caption{An example of discriminant label information with missing labels. If we set the rank of the label matrix to $3$ or $4$, then the missing value of the label matrix will be correspondingly set to $1$ or $-1$. Obviously, the label matrix with larger rank provides more discriminant information.   }
	\label{discriminant_ls}
\end{figure}

Another is label discriminant information. Since label matrix has missing entries, it is quite difficult to obtain the true rank of the label matrix in advance. If the rank is set too small, it will inevitably lose useful label discriminant information, hence affecting unfavorably the classification performance. An example of this phenomenon is shown in Fig. \ref{discriminant_ls}. Therefore, only if the rank of the label matrix is large enough, can the recovered label matrix provide more discriminant information.

Nevertheless, to the best of our knowledge, current methods have not considered incorporating local low-rank label structure and label discriminant information into a single low-rank model to improve performance. Therefore, how to embed both label structures into the low-rank model elegantly and effectively is still very challenging at present. To achieve this, we develop a \textbf{D}iscriminative \textbf{M}ulti-\textbf{L}abel \textbf{L}earning (DM2L) model for multi-label learning with missing labels. Our model is not only simple and effective, but also capable of jointly capturing global low-rank label structure, local low-rank label structure and label discrimination information in the original label space, which is supported by theoretical analysis. Besides, we provide a nonlinear extension of DM2L by kernel trick to
enhance its ability and develop a concave-convex programming to solve
these optimization objectives. Compared to the other low-rank based methods, our method involves the fewest assumptions and only one hyper-parameter. Even so, our method still outperforms the state-of-the-art methods as shown in the experiments.

The rest of the paper is organized as follows. In Section 2, we review some related works. In Section 3, the proposed framework is presented. Experimental results are presented in Section 4. Finally, Section 5 gives some concluding remarks.

\section{Related Work}

In multi-label classification, valid labels are not throughly provided by the training set. This problem is referred to as the presence of missing labels. The common approaches to tackle this problem can roughly be divided as:

(1) Pre-processing methods, which attempt to recover the label matrix as a pre-processing and then train mutli-label classifiers with the complete labels. For example,  MLML-exact \cite{wu2014multi} and Ml-mg \cite{wu2015ml} conduct label completion based on label consistency and label smoothness; LSR \cite{lin2013image} carries out image tag completion via image-specific and tag-specific linear sparse reconstructions; FastTag \cite{chen2013fast} assumes that the observed incomplete labels can be linearly transformed to the unobserved complete ones and learns this linear mapping via the idea of training a denoising auto-encoder. These methods do not integrate the consequences of multi-label classification on label recovery, which may limit their effectiveness.

(2) Transductive methods, which attempt to recovery the label matrix in the transductive learning setting. For example, MC-1 \cite{cabral2011matrix, goldberg2010transduction} and IrMMC \cite{liu2015low} first make a matrix containing all the feature vectors and the label vectors of both the training and the test data. Then, they exploit matrix completion methods to fill in the missing entries of this matrix. These methods do not have inductive capabilities, which limits their scopes of application.

\begin{figure}[htbp]
	\setlength{\abovecaptionskip}{0.2cm}
	\setlength{\belowcaptionskip}{0.cm}
	\centering
	\includegraphics[width=0.8\textwidth]{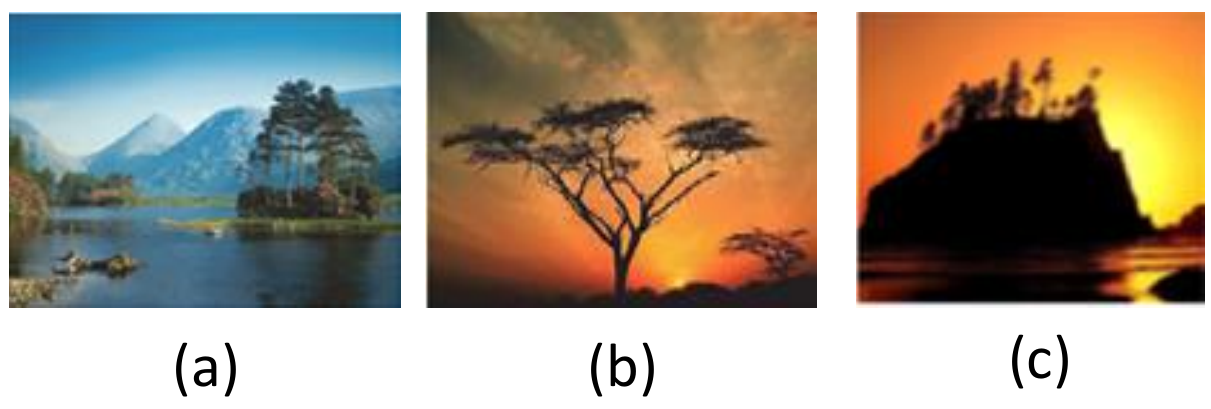}
	\caption{An example of the limitation of traditional assumption for local label structure. According to the assumption that the samples with similar features should have similar labels, the similarity between picture a and c is less than that between b and c. However, from the perspective of a particular label 'River', the similarity between a and c should be greater than that between b and c.}
	\label{local_structure}
\end{figure}

(3) Synchronized methods, which attempt to learn the multi-label classifiers and recover the label matrix simultaneously. LARS\cite{yuan2006model} uses group lasso to selectively penalize the pairwise ranking errors between the two partitions. MLR-GL\cite{bucak2011multi} uses a ranking based multi-label learning framework to handle missing labels. MLMLFS \cite{zhu2018multi} recovers missing labels by the robust linear regression. SMILE \cite{tan2017semi} uses a pre-defined label correlation matrix to estimate the likelihood of missing labels. SVMMN \cite{liu2018svm} integrates both sample smoothness and class smoothness into criterion function to solve the missing label problem. LSML \cite{huang2019improving} attempts to learn a label correlations matrix, which can be exploited to augment the incomplete label matrix and obtain a new supplementary label matrix. However, these methods do not exploit low-rank label stuctures, which is useful to recovery missing labels. To achieve this, both LEML \cite{yu2014large} and REML \cite{xu2016robust} learn a linear low-rank instance-to-label mapping. ML-LRC \cite{xu2014learning} incorporates a low-rank supplementary label matrix to augment the possibly incomplete label matrix. fPML \cite{yu2018feature} decomposes label matrix into two low-rank matrices. From the perspective of matrix factorization, the product of the two low-rank matrices of fPML gives the low-rank approximation of the original label matrix. iMVWL \cite{tan2018incomplete} learns a low-rank label correlation matrix and multiplies this correlation matrix with original label matrix to replenish missing labels. Unfortunately, these low-rankness exploiting methods only capture global low-rank label structure, but ignore local label structure. To address this issue, GLOCAL first uses $K$-means clustering algorithm to partition dataset into several local groups, then exploits global and local label structures simultaneously, through learning a latent label representation and optimizing label manifolds. Although GLOCAL attempts to jointly explore local and global label structures, it is too complicated because of involving many regularization terms and thus hyper-parameters. Moreover, GLOCAL does not attempt to model the more essential local low-rank label structure, but explores a hypothetical local label structure. Thus, it may be limited by the assumption. Specifically, to explore the local label structure, it assumes that the samples with similar features should have similar labels. However, a label is usually determined by a subset of features of an instance. As a result, although two instances own very similar features, they may behave differently in a particular label, as illustrated in Fig. \ref{local_structure}.

\section{The Proposed Method}
In this section, we introduce the proposed method - called Discriminant Multi-label Learning with Missing Labels (DM2L). Section 3.1 presents the building ideas of DM2L. In Section 3.2 we build an interesting nonlinear extension of DM2L. Subsequently, we detail how to optimize DM2L in Section 3.3. Finally, the time complexity analysis is given in Section 3.4.
\subsection{Discriminant Multi-label Learning with Missing Labels (DM2L)}
In the context of multi-label learning, let matrix $\mathbf{X}=[\mathbf{x}_1, \cdots, \mathbf{x}_n]^T \in \mathbb{R}^{n \times d}$ and $\mathbf{Y}=[\mathbf{y}_1, \cdots, \mathbf{y}_n]^T \in \{-1, 1\}^{n \times c}$ refer to the instance matrix and the true label matrix, respectively, where $n \geq d$, $d$ is the feature dimension of an instance, $n$ is the number of instances and $c$ is the size of label set. Since  the label matrix is generally incomplete in the real-world applications, we assume $\widetilde{\mathbf{Y}}  \in \mathbb{R}^{n \times c}$ to be the observed label matrix, where many entries are unknown. Let $\boldsymbol{\Omega} \subseteq \{1, \cdots, n\} \times \{1, \cdots, c\}$ denote the set of the indices of the observed entries in $\mathbf{Y}$, we can define a linear operator $\mathcal{R}_{\boldsymbol{\Omega}}(\mathbf{Y}): \mathbb{R}^{n \times c} \mapsto  \mathbb{R}^{n \times c}$ as

\begin{equation}
\widetilde{\mathbf{Y}}_{i, j}=[\mathcal{R}_{\boldsymbol{\boldsymbol{\Omega}}}(\mathbf{Y})]_{i,j}=\left\{\begin{split}
\mathbf{Y}_{i,j} \ (i, j) \in \boldsymbol{\Omega}\\
0 \  \ (i, j) \notin \boldsymbol{\Omega}
\end{split}\right.
\end{equation}

Following the common assumption of low-rank for $\mathbf{Y}$  as in \cite{yang2016improving,xu2014learning}, we can assume the rank of $\mathbf{Y}$ is $r$ ($ n \geq c \geq r$) without loss of generality. Then, the multi-label learning problem becomes : given $\widetilde{\mathbf{Y}}$ and $\mathbf{X}$, how to find the optimal $\mathbf{W} \in \mathbb{R}^{d \times c}$ so that the estimated label matrix $\mathbf{XW}$ can be as close to the ground-truth label matrix $\mathbf{Y}$ as possible. To this end, we can make use of the low-rank property of $\mathbf{Y}$ and optimize the following objective function:

\begin{equation}
\label{ob1}
\min_{\mathbf{W}} \frac{1}{2}\|\mathcal{R}_{\boldsymbol{\Omega}}(\mathbf{XW}) - \tilde{\mathbf{Y}}\|_F^2 + \lambda_d \|\mathbf{XW}\|_*
\end{equation}
where $\|\cdot\|_*$ denotes the nuclear-norm of a matrix, which is commonly used to encourage low-rank structure of a matrix.

However, just forcing the low rankness for $\|\mathbf{XW}\|_*$ is not sufficient, since in addition to this observation, the label matrix still has some other intrinsic properties below, \begin{enumerate}
	\item If we denote $\mathbf{X}_k$ as the set of training instances associated with label $k$ and $\mathbf{Y}_k$ as the label matrix corresponding to $\mathbf{X}_k$, then we can easily see that the $k$-th column of $\mathbf{Y}_k$ is a vector with all elements to be $1$ and $\operatorname{rank}(\mathbf{Y}_k) \leq c$.
	\item If label $k$ and label $q$ have high possibilities to (asymmetrically) co-occur, or the two labels never appear at the same time, then the $q$-th column of $\mathbf{Y}_k$ is a vector with all elements to be $1$ or $-1$. Thus label $k$ and label $q$ are linearly dependent (in linear algebra sense) and $\operatorname{rank}(\mathbf{Y}_k) \leq c-1$.
	\item Because most labels are unnecessarily linearly dependent, the rank of $\mathbf{XW}$ should be made as high as possible in the case both the local and global structures are maintained to ensure that the learned model has strong discriminability.
\end{enumerate}

In order to take advantage of the above-analyzed label structures,  we impose, on the one hand, low-rank structures for the whole predictions of instances from the same labels, and on the other hand, a maximally separated structure for all the predictions of instances from different labels. More precisely, two new nuclear-norm-based terms is introduced to replace the original nuclear norm in the objective function (\ref{ob1}), i.e.,

\begin{equation}
\label{ob3}
\begin{split}
\min_{\mathbf{W}} \frac{1}{2}\|\mathcal{R}_{\boldsymbol{\Omega}}(\mathbf{XW}) - \tilde{\mathbf{Y}}\|_F^2 +\lambda_d \left(\sum_{k=1}^c \|\mathbf{X}_k\mathbf{W}\|_*-\|\mathbf{XW}\|_*\right)
\end{split}
\end{equation}
where $\lambda_d$ is a hyper-parameter trading off the two terms, the term $\|\mathbf{X}_k\mathbf{W}\|_*$ encourages a low-rankness for the predictions of instances from the label $k$ to capture the local label structure in label $k$, and the term $-\|\mathbf{XW}\|_*$ encourages a maximally separated structure for the predictions of instances from different labels to provide more underlying discriminant information because two linearly independent predicting vectors correspond to two different label vectors, such an independence leads to desirable discrimination among different labels. Moreover, according to Theorem \ref{tho1} (proved in Appendix A.), the term $\sum_{k=1}^c \|\mathbf{X}_k\mathbf{W}\|_*$ is an upper bound of $\|\mathbf{XW}\|_*$ \footnote{Note that $\sum_{k=1}^c \operatorname{rank}\left (\mathbf{X}_k\mathbf{W}\right) \geq  \operatorname{rank}\left(\mathbf{XW}\right)$ is an obvious conclusion, which can be found in the linear algebra textbook \cite{strang1993introduction}. However, the proposition about $\sum_{k=1}^c \|\mathbf{X}_k\mathbf{W}\|_* \geq \|\mathbf{XW}\|_*$ needs to be proved particularly, because the proof is no longer trivial in this case.}, implying that the global low-rank label structure can be also maintained with a suitable $\lambda_d$.

%


\begin{theorem}
	\label{tho1}
	Let $\mathbf{A, B}$ and $\mathbf{C}$ be matrices of the same row dimensions, and $[\mathbf{A, C}]$ be concatenation of $\mathbf{A}$ and $\mathbf{C}$, Likewise for $[\mathbf{A, B, C}]$ and $[\mathbf{B, C}]$. Then, we have
	\begin{equation}
	\|[\mathbf{A, B, C}]\|_* \leq \|[\mathbf{A, C}]\|_*+ \|[\mathbf{B, C}]\|_*.
	\end{equation}
\end{theorem}

\begin{remark}
	As can be seen from the objective function (\ref{ob3}), our method only assumes that the label matrix is low-rank, so this method is one of the low-rank methods involving the fewest assumptions. As for the local low-rank label structures and label discriminant information, both are structure information inherent in multi-label learning, and thus do not involve any additional assumptions.
\end{remark}

\subsection{Nonlinear Extension}

So far, we have only considered linear model, thus limiting the ability to characterize feature interactions. Deep learning \cite{lezama2018ole} or kernel methods can be used to achieve this purpose. Here, we directly adopt the kernel trick to build an interesting nonlinear extension of DM2L, because DM2L can be recast into an equivalent dual representation as shown in Theorem \ref{representer2} (proved in Appendix B), in which the predictions are based on linear combinations of a \emph{kernel function} evaluated at the training data points.

\begin{theorem}
	\label{representer2}
	Let $\kappa$ be a kernel on $\mathcal{X}$, $\phi(\mathbf{x})$ be its associate nonlinear feature space mapping, and $f(\mathbf{X})=[\phi(\mathbf{x}_1), \cdots, \phi(\mathbf{x}_n)]^T$, the optimal solution $\mathbf{W}^*=[\mathbf{w}_1^*, \cdots, \mathbf{w}_c^*]$ of the optimization problem
	\begin{equation}
	\label{kernelObj}
	\begin{split}
	\min_{\mathbf{W}} \frac{1}{2}&\|\mathcal{R}_\Omega \left(f(\mathbf{X})\mathbf{W}\right) - \tilde{\mathbf{Y}}\|_F^2 \\&+ \lambda_d \left(\sum_{k=1}^c \|\left(f(\mathbf{X_k}\right)\mathbf{W}\|_*-\|f(\mathbf{X})\mathbf{W}\|_*\right)
	\end{split}
	\end{equation}
	can be expressed as:
	\begin{equation}
	\label{repre3}
	\mathbf{w}_j^*=\sum_{i=1}^{n} a_{i}^j \phi(\mathbf{x}_{i})
	\end{equation}
	and
	\begin{equation}
	\label{total_repre3}
	\mathbf{W}^{*}=\left[\phi\left(\mathbf{x}_{1}\right), \phi\left(\mathbf{x}_{2}\right), \cdots, \phi\left(\mathbf{x}_{n}\right)\right] \mathbf{A}
	\end{equation}
	where $\mathbf{w}_j$ is the $j$-th column vector of $\mathbf{W}$ and $\mathbf{a}_j=[a^j_1, \cdots, a^j_n]^T$ is the $j$-th column vector of $\mathbf{A}= [\mathbf{a}_1, \cdots, \mathbf{a}_c]$.
	
\end{theorem}

By plugging formulas (\ref{repre3}) and (\ref{total_repre3}) into (\ref{kernelObj}), it can be shown that the optimization problem (\ref{kernelObj}) is equivalent to problem (\ref{ob4}):
\begin{equation}
\label{ob4}
\begin{split}
\min_{\mathbf{A}} &\frac{1}{2}\|\mathcal{R}_\Omega(\mathbf{KA}) - \tilde{\mathbf{Y}}\|_F^2 + \lambda_d \left(\sum_{k=1}^c \|\mathbf{K}_k\mathbf{A}\|_*-\|\mathbf{KA}\|_*\right)
\end{split}
\end{equation}
where $\mathbf{K}=f(\mathbf{X})f(\mathbf{X})^T$, $\mathbf{K}_k=f(\mathbf{X}_k)f(\mathbf{X})^T$ and $f(\mathbf{X})=[\phi(\mathbf{x}_1), \cdots, \phi(\mathbf{x}_n)]^T$.

Actually, if we use a linear kernel, problem (\ref{ob4}) is reduced to (\ref{ob3}).
To sum up, we can draw the following conclusions from the objective function (\ref{ob4}):
\begin{enumerate}
	\item This method is one of the low-rank methods involving the least assumptions, thus it only involves one hyper-parameter and much simpler than existing ones.
	\item The label structures are directly explored in the \emph{original} label space.
	\item The term $\sum_{k=1}^c \|\mathbf{K}_k\mathbf{A}\|_*$ in (\ref{ob4}) is able to capture both local and global low-rank label structures.
	\item The term $-\|\mathbf{KA}\|_*$ can provide more underlying discriminant information.
	\item According to Theorem \ref{tho1}, the value of (\ref{ob4}) is always greater than $0$, thus (\ref{ob4}) will not degenerate for any $\mathbf{A}$.
	\item Local label structure explored in (\ref{ob3}) and (\ref{ob4}) has a significant advantage over existing methods \cite{huang2012multi,zhu2018multi}. As stated in the introduction, existing methods usually assume that samples having similar features have similar labels, which is always not true for a particular label. Instead, (\ref{ob3}) and (\ref{ob4}) is not built on this assumption, so there is no such a limitation.
\end{enumerate}

\subsection{The Concave-convex Programming}
Note if we use a linear kernel, problem (\ref{ob4}) is reduced to (\ref{ob3}), thus we just focus on optimizing problem (\ref{ob4}). Because the problem (\ref{ob4}) is non-differentiable and non-convex, the traditional convex optimization methods, e.g., gradient descent, Newton method, are not suitable for this problem. However, our objective function $\mathcal{J}(\mathbf{A})$ can be rewritten as the sum of a convex part $\mathcal{J}_{vex}(\mathbf{A})$ and a concave part $\mathcal{J}_{cave}(\mathbf{A})$, i.e.,
\begin{equation}
\begin{split}
\mathcal{J}(\mathbf{A})=&\mathcal{J}_{vex}(\mathbf{A})+\mathcal{J}_{cave}(\mathbf{A})\\
=&\left[\frac{1}{2}\|\mathcal{R}_\Omega(\mathbf{KA}) - \tilde{\mathbf{Y}}\|_F^2 + \lambda_d \sum_{k=1}^c \|\mathbf{K}_k\mathbf{A}\|_*\right]\\ &+\left[-\lambda_d \|\mathbf{KA}\|_*\right].
\end{split}
\end{equation}
Therefore, the problem is a D.C. (difference of convex functions) program and also similar to the D.C. program in \cite{qiu2015learning}. We likewise use the concave-convex programming (CCCP) to solve problem (\ref{ob4}) as in \cite{qiu2015learning}.

As we know, CCCP is a special case of the majorisation-minimisation (MM) algorithm, which is a surrogate type optimization method. Algorithm 1 illustrate the update procedure of the CCCP algorithm, where at each iteration we firstly use a surrogate objective $\mathcal{J}_t(\mathbf{A})$ that majorises the original objective at the current solution $\mathbf{A}_t$, then apply any optimization algorithm to the surrogate for the next update $\mathbf{A}_{t+1}$.

In Algorithm 1, we need to determine what surrogate function to use. Due to that the next updates $\mathbf{A}_{t+1}$ can be obtained by matching the convex part subgradient to the concave part subgradient:
\begin{equation}
\mathbf{A}_{t+1} \  satisfies \ \partial \mathcal{J}_{vex}(\mathbf{A}_{t+1}) = -\partial \mathcal{J}_{cave}(\mathbf{A}_{t}),
\end{equation}
where $\partial \mathcal{J}(\mathbf{A})$ is a subgradient of $\mathcal{J}(\mathbf{A})$, we use the convex term $\mathcal{J}_{vex}(\mathbf{A})$ plus the linear term $trace\left(\partial \mathcal{J}_{cave}(\mathbf{A}_{t})\mathbf{A}'\right)$ as the surrogate objective function $\mathcal{J}_t(\mathbf{A})$.

Moreover, in Algorithm 1, we also need to compute the subgradient of nuclear norm $\|\cdot\|_*$, which can be evaluated using the simple approach shown in Algorithm 2 \cite{watson1992characterization,lezama2018ole}.

\begin{algorithm}[htb!]
	\small
	\caption{The concave-convex programming}
	\label{alg1}
	\begin{algorithmic}[1]
		\REQUIRE
		Training dataset matrix $\mathbf{X}$ and the hyperparameter $\lambda_d$.
		\ENSURE
		The learned optimal $\mathbf{A}_*$.
		\STATE Initialize $\mathbf{A}_0$ with the identity matrix.
		\REPEAT
		\STATE \begin{equation}
		\footnotesize
		\begin{aligned} \mathbf{A}_{t+1}=& \arg \min _{\mathbf{A}} \mathcal{J}_{v e x}(\mathbf{A})+\operatorname{trace}\left(\partial \mathcal{J}_{c a v e}\left(\mathbf{A}_{t}\right) \mathbf{A}^{\prime}\right) \\=& \arg \min _{\mathbf{A}} \frac{1}{2}\|\mathcal{R}_\Omega(\mathbf{KA}) - \tilde{\mathbf{Y}}\|_F^2 + \\ & \lambda_d \sum_{k=1}^c \|\mathbf{K}_k\mathbf{A}\|_*-\lambda_{d} \operatorname{trace}\left(\mathbf{K}^{\prime} \partial\left\|\mathbf{K} \mathbf{A}_{t}\right\|_{*} \mathbf{A}^{\prime}\right) \end{aligned}
		\end{equation}
		\UNTIL{convergence or stopping criteria}
	\end{algorithmic}
\end{algorithm}

\begin{algorithm}[htb!]
	\small
	\caption{An approach to evaluate a subgradient of matrix nuclear norm}
	\label{alg2}
	\begin{algorithmic}[1]
		\REQUIRE
		A $n \times c$ matrix $\mathbf{C}$ and a small threshold value $\delta$.
		\ENSURE
		A subgradient of the nuclear norm $\partial \|\mathbf{C}\|_*$.
		\STATE Perform singular value decomposition: $$\mathbf{C=U\Sigma V}$$
		\STATE $s \leftarrow$ the number of singular values larger than $\delta$.
		\STATE Partition $\mathbf{U}$ and $\mathbf{V}$ as
		$$\mathbf{U}=[\mathbf{U}^1, \mathbf{U}^2]$$
		$$\mathbf{V}=[\mathbf{V}^1, \mathbf{V}^2]$$
		where $\mathbf{U}^1$ and $\mathbf{V}^1$ have $s$ columns;
		\STATE $\partial \|\mathbf{C}\|_* \leftarrow \mathbf{U}^1 {\mathbf{V}^1}'$;
	\end{algorithmic}
\end{algorithm}

\subsection{Time Complexity}{\label{timeCom}}

In our proposed methods, $\mathbf{X} \in \mathbb{R}^{n \times d}$, $\mathbf{Y} \in \{-1, 1\}^{n \times c}$,  $\mathbf{W} \in \mathbb{R}^{d \times c}$, $\mathbf{K} \in \mathbb{R}^{n \times n}$ and $\mathbf{A} \in \mathbb{R}^{n \times c}$, where $n$ is the number of instances, $d$ is the feature dimensionality and $c$ is the number of labels. The time complexities of Algorithm \ref{alg1} and Algorithm 2 are dominated by matrix multiplication and SVD operations. Because we provide both linear and nonlinear versions of DM2L, their time complexities are different.
		
For the linear version of DM2L (denoted as DM2L-l), where the goal is optimizing $\mathbf{W}$, the calculation of a subgradient of $\mathbf{C} \in \mathbb{R}^{n \times c}$ in Algorithm 2 leads to a complexity of $\mathcal{O}(n^2c+c^3)$. In each iteration of Algorithm \ref{alg1}, the calculating subgradient of  $\mathcal{J}(\mathbf{W})$ leads to a time complexity of $\mathcal{O}(n^2c+c^3 + ndc )$. Therefore, the total complexity of DM2L-l is $ \mathcal{O}\left(\gamma (n^2c+c^3 + ndc )\right)$, where $\gamma$ is the number of iterations.
		
For the nonlinear version of DM2L (denoted as DM2L-nl), where the goal is optimizing $\mathbf{A}$, the calculation of a subgradient of $\mathbf{C} \in \mathbb{R}^{n \times c}$ in Algorithm 2 leads to a complexity of $\mathcal{O}(n^2c+c^3)$. In each iteration of Algorithm \ref{alg1}, the calculating subgradient of  $\mathcal{J}(\mathbf{A})$ leads to a time complexity of $\mathcal{O}(n^2c+c^3)$. Therefore, the total complexity of DM2L-nl is $ \mathcal{O}(\gamma n^2c+\gamma  c^3)$, where $\gamma$ is the number of iterations.

\section{Experiments}
In this section, we conduct extensive experiments to testify the performance of DM2L-l and DM2L-nl on both the full-label case and the missing label case.

\subsection{Datasets}
Extensive experiments are performed on text and image datasets. Specifically, on text,  four Yahoo datasets (Arts, Business, Recreation, Social) \footnote{http://lamda.nju.edu.cn/files/MDDM-expdata.rar}, the Tmc2007\footnote{http://mulan.sourceforge.net/data sets-mlc.html \label{ds}} dataset and enron\textsuperscript{\ref {ds}} dataset are used. On images, the Corel16k \textsuperscript{\ref {ds}}  and Image \footnote{http://cse.seu.edu.cn/people/zhangml/files/Image.rar} datasets are used. In the sequel, we denote each dataset by its first three letters. For each dataset, we randomly select 60 percent of the instances for training, and the rest for testing. To reduce statistical variability, results are averaged over $30$ independent repetitions.

\begin{table}[htp!]
	\centering
	\small
	\caption{A summary of the attributes of each dataset.}
	\label{dataset}
	\begin{tabular}{lcccc}
		\hline \hline
		Dataset       & \#instance   & \#dim      & \#label    & \#label/instance   \\ \hline
		Arts          & 5,000 & 462   & 26  & 1.64 \\
		Business      & 5,000 & 438   & 30  & 1.59 \\
		Recreation    & 5,000 & 606   & 22  & 1.42 \\
		Social        & 5,000 & 1,047 & 39  & 1.28 \\
		Enron         & 1,702 & 1,001 & 53  & 3.37 \\
		Image         & 2,000 & 294   & 5   & 1.24 \\ 
		{{Corel16k}}      &{{13,811}} & {{500}}   & {{161}} & {{2.87}} \\
		{{Tmc2007-500}}    &{{28,596}} & {{500}}   & {{22}}  & {{2.16}} \\ \hline \hline
	\end{tabular}
\end{table}

\subsection{Performance evaluation}
We use four popular metrics in multi-label learning \cite{zhu2018multi}, i.e., Ranking loss (Rkl), Average Area Under the ROC Curve (AUC), Coverage (Cvg) and Average precision (Ap). For Auc and Ap, the higher value the better; whereas for Rkl and Cvg, the lower value the better.
We use four popular metrics in multi-label learning \cite{zhu2018multi} as below:

Let $p$ be the number of test instances, $\mathbf{C}_i^+$, $\mathbf{C}_i^-$ be the sets of positive and negative labels associated with the $i$th instance; and $\mathbf{Z}_j^+$, $\mathbf{Z}_j^-$ be the sets of positive and negative instances belonging to the $j$th label. Given input $\mathbf{x}$, let rank$_f(\mathbf{x}, y)$ be the rank of label $y$ in the predicted label ranking (sorted in descending order).

\begin{itemize}
	\item Ranking loss (Rkl): This is the fraction that a negative label is ranked higher than a positive label. For test instance $\mathbf{x}_i$, define $\mathbf{Q}_i = \{(j', j'') | f_{j'}(\mathbf{x}_i) \leq f_{j''}(\mathbf{x}_i), (j', j'') \in \mathbf{C_i}^+ \times \mathbf{C}_i^-\}$. Then, Rkl$ = \frac{1}{p} \sum_{i=1}^p \frac{|\mathbf{Q}_i|}{|\mathbf{C_i}^+| |\mathbf{C_i}^-|}$.
	\item Average Area Under the ROC Curve (AUC): This is the fraction that a positive instance is ranked higher than a negative instance, averaged over all labels. Specifically, for label $j$, define $\mathbf{\tilde{Q}}_j=\{(i', i'')|f_j(\mathbf{x}_{i'}) \geq f_j(\mathbf{x}_{i''}), (\mathbf{x}_{i'}, \mathbf{x}_{i''}) \in  \mathbf{Z}_j^+ \times \mathbf{Z}_j^-\}$. Then, Auc = $\frac{1}{c} \sum_{j=1}^c \frac{|\mathbf{\tilde{Q}}_j|}{| \mathbf{Z}_j^+|| \mathbf{Z}_j^-|}$.
	\item Coverage (Cvg): This counts how many steps are needed to move down the predicted label ranking so as to cover all the positive labels of the instances. $Cvg = \frac{1}{p}\sum_{i=1}^p \max \{rank_f (\mathbf{x}_i, j)|j \in \mathbf{C}_i^+\}-1$.
	\item Average precision (Ap): This is the average fraction of positive labels ranked higher than a particular positive label. For instance $\mathbf{x}_i$, define $\mathbf{\tilde{Q}}_{i,c}=\{j | rank_f (\mathbf{x}_i, j) \leq rank_f (\mathbf{x}_i, c), j \in \mathbf{C}_i^+\}$. Then, Ap$=\frac{1}{p} \sum_{i=1}^p \frac{1}{|\mathbf{C_i}^+|} \sum_{c \in \mathbf{C_i}^+ \frac{|\mathbf{\tilde{Q}}_{i,c}|}{rank_f (\mathbf{x}_i, c)}}$.
\end{itemize}

\subsection{Comparison methods}
We compare DM2L \footnote{The source code is available at "https://github.com/John986/Multi-label-Learning-with-Missing-Labels"} to the following state-of-the-art multi-label learning algorithms:
\begin{itemize}
	\item Low-rank empirical risk minimization for multi-label learning (LEML) \cite{yu2014large} learns a linear instance-to-label mapping with low-rank structure to take advantage of global label structure. {The rank $k$ is tuned in $\{5, 10, 15, \dots\}$ and the regularization parameter is tuned in $\{10^{-5}, 10^{-4}, \cdots, 10^{1}\}$.}
	\item Learning low-rank label correlations for multi-label classification (ML-LRC) \cite{xu2014learning} is an integrated framework, which learns the correlations among labels while training the multi-label model simultaneously.  {Parameters $\lambda_1$, $\lambda_2$ and $\lambda_3$ are selected from $\{10^{-5}, 10^{-4}, \cdots, 10^{0}\}$. For simplity, when one parameter is tuned, the others are fixed to their best.}
	\item Multi-label learning with global and local label correlation (GLOCAL) \cite{zhu2018multi} learns and exploits global and local label structures for multi-label learning with missing labels. {Parameters $\lambda =1$, $\lambda_1$ to $\lambda_5$ are selected from $\{10^{-5}, 10^{-4}, \cdots, 10^{1}\}$, $k$ is tuned in $\{5, 10, 15, 20\}$. For simplity, when one parameter is tuned, the others are fixed to their best.}
	\item {Label-specific features for multi-label classification with missing labels (LSML)  \cite{huang2019improving} performs classification and label matrix recovery jointly. All the parameters are tuned in $\{10^{-5}, 10^{-4}, \cdots, 10^{3}\}$.}
\end{itemize}

The hyper-parameters in compared methods are selected via 5-fold cross-validation on the training set. {{The parameter $\lambda_d$ of DM2L-l and DM2L-nl is selected from $\{10^{-5}, 10^{-4},\cdots, 10^5\}$.  The threshold value $\delta$ is fixed at $0.005$ in Algorithm (2), which performs consistently well in our experiments.}} For DM2L-nl, the gaussian kernel function is used in our experiments. The formulation of gaussian kernel function is shown below:
\begin{equation}
\label{gaussianF}
K\left(\mathbf{x}, \mathbf{x}^{\prime}\right)=\exp \left(-\frac{\left\|\mathbf{x}-\mathbf{x}^{\prime}\right\|_{2}^{2}}{2 \sigma^{2}}\right),
\end{equation}
{{and its parameter $\sigma$ is chosen from $\{0.5, 1, 1.5, 2\}$.}}

\subsection{Experimental results}

{{In this subsection, we apply DM2L-l and DM2L-nl to the multi-label learning with both full labels and missing labels. To generate missing labels, we randomly sample $\rho$ of the elements in the label matrix as observed, and the rest as missing. When $\rho=1$, it reduces to the full-label case. Table \ref{full_result}, Table \ref{miss7_result} and Table \ref{miss3_result} show the label prediction results on the test data with full labels ($\rho=1$) and missing labels ($\rho=0.7$ and $\rho=0.3$), respectively. From the persective of observation, DM2L-nl and DM2L-l are better than the other compared methods in general. In particular, DM2L-nl performs best nearly on all evaluation measures and datasets. }}

\begin{table*}[htp!]
	\caption{Results for learning with full labels ($\rho = 1$).}
	\label{full_result}
	\centering
	\scriptsize
	\begin{tabular}{ccccccccc}
		\hline \hline
		&  & LEML   & ML-LRC & GLOCAL & LSML   & DM2L-l & DM2L-nl\\
		\hline
		\multirow{4}{*}{Art} &	Rkl	&	0.167	&	0.161	&	0.152	&	0.127	&	0.139	&	\textbf{0.125}	\\
		&	Auc	&	0.835	&	0.841	&	0.85	&	\textbf{0.876}	&	0.864	&	0.875	\\
		&	Cvg	&	6.305	&	5.636	&	5.852	&	4.978	&	5.409	&	\textbf{4.965}	\\
		&	Ap	&	0.596	&	0.484	&	0.61	&	\textbf{0.625}	&	0.623	&	0.61	\\ \hline
		\multirow{4}{*}{Bus} &	Rkl	&	0.056	&	0.04	&	0.047	&	0.043	&	0.046	&	\textbf{0.039}	\\
		&	Auc	&	0.945	&	\textbf{0.962}	&	0.954	&	0.958	&	0.956	&	0.961	\\
		&	Cvg	&	3.153	&	\textbf{2.333}	&	2.656	&	2.521	&	2.674	&	2.419	\\
		&	Ap	&	0.864	&	0.885	&	0.877	&	0.881	&	0.882	&	\textbf{0.887}	\\ \hline
		\multirow{4}{*}{Rec}&	Rkl	&	0.1788	&	0.149	&	0.159	&	0.152	&	0.147	&	\textbf{0.136}	\\
		&	Auc	&	0.8252	&	0.855	&	0.845	&	0.853	&	0.857	&	\textbf{0.864}	\\
		&	Cvg	&	5.0164	&	4.082	&	4.542	&	4.365	&	4.2	&	\textbf{3.983}	\\
		&	Ap	&	0.6027	&	0.575	&	0.618	&	0.631	&	\textbf{0.635}	&	0.633	\\ \hline
		\multirow{4}{*}{Enr}&	Rkl	&	0.172	&	0.121	&	0.117	&	0.136	&	0.131	&	\textbf{0.109}	\\
		&	Auc	&	0.83	&	0.882	&	0.885	&	0.866	&	0.871	&	\textbf{0.892}	\\
		&	Cvg	&	20.37	&	15.477	&	\textbf{15.430}	&	18.346	&	17.714	&	15.718	\\
		&	Ap	&	0.589	&	0.603	&	0.632	&	0.634	&	0.588	&	\textbf{0.648}	\\ \hline
		\multirow{4}{*}{Ima}&	Rkl	&	0.203	&	0.182	&	0.18	&	0.181	&	0.193	&	\textbf{0.138}	\\
		&	Auc	&	0.797	&	0.819	&	0.82	&	0.819	&	0.807	&	\textbf{0.862}	\\
		&	Cvg	&	1.069	&	0.996	&	0.992	&	0.993	&	1.044	&	\textbf{0.827}	\\
		&	Ap	&	0.758	&	0.781	&	0.783	&	0.783	&	0.76	&	\textbf{0.833}	\\ \hline
		\multirow{4}{*}{Soc}&	Rkl	&	0.106	&	0.08	&	0.078	&	0.062	&	0.065	&	\textbf{0.053}	\\
		&	Auc	&	0.894	&	0.92	&	0.922	&	0.938	&	0.935	&	\textbf{0.947}	\\
		&	Cvg	&	5.62	&	4.09	&	4.012	&	3.544	&	3.711	&	\textbf{3.077}	\\
		&	Ap	&	0.723	&	0.596	&	0.666	&	0.777	&	0.78	&	\textbf{0.783}	\\ \hline
		\multirow{4}{*}{Cor}&	Rkl	&	0.164	&	0.152	&	0.175	&	\textbf{0.142}	&	0.146	&	0.143	\\
		&	Auc	&	0.836	&	0.848	&	0.825	&	\textbf{0.860}	&	0.854	&	0.857	\\
		&	Cvg	&	47.896	&	45.377	&	49.909	&	\textbf{42.811}	&	43.673	&	43.323	\\
		&	Ap	&	0.328	&	0.293	&	0.306	&	0.335	&	0.328	&	\textbf{0.339}	\\ \hline
		\multirow{4}{*}{Tmc}&	Rkl	&	0.046	&	0.044	&	0.046	&	0.047	&	0.046	&	\textbf{0.030}	\\
		&	Auc	&	0.954	&	0.956	&	0.954	&	0.954	&	0.955	&	\textbf{0.970}	\\
		&	Cvg	&	2.874	&	2.701	&	2.855	&	2.88	&	2.838	&	\textbf{2.273}	\\
		&	Ap	&	0.833	&	0.821	&	0.833	&	0.833	&	0.833	&	\textbf{0.886}	\\
		\hline  \hline
	\end{tabular}
\end{table*}

\begin{table*}[htp!]
	\caption{Results for learning with missing labels ($\rho = 0.7$).}
	\label{miss7_result}
	\centering
	\scriptsize
	\begin{tabular}{ccccccccc}
		\hline \hline
		&  & LEML   & ML-LRC & GLOCAL & LSML   & DM2L-l & DM2L-nl\\
		\hline
		\multirow{4}{*}{Art}&	Rkl	&	0.199	&	0.164	&	0.16	&	0.145	&	0.138	&	\textbf{0.119}	\\
		&	Auc	&	0.804	&	0.839	&	0.843	&	0.857	&	0.864	&	\textbf{0.881}	\\
		&	Cvg	&	7.212	&	5.711	&	6.079	&	5.617	&	5.382	&	\textbf{4.810}	\\
		&	Ap	&	0.556	&	0.479	&	0.6	&	0.608	&	0.621	&	\textbf{0.623}	\\ \hline
		\multirow{4}{*}{Bus}&	Rkl	&	0.072	&	0.048	&	0.045	&	0.054	&	0.043	&	\textbf{0.037}	\\
		&	Auc	&	0.93	&	0.953	&	0.957	&	0.947	&	0.958	&	\textbf{0.963}	\\
		&	Cvg	&	3.905	&	2.8	&	2.532	&	3.097	&	2.571	&	\textbf{2.317}	\\
		&	Ap	&	0.835	&	0.876	&	0.877	&	0.867	&	0.882	&	\textbf{0.889}	\\ \hline
		\multirow{4}{*}{Rec}&	Rkl	&	0.211	&	0.15	&	0.161	&	0.177	&	0.149	&	\textbf{0.131}	\\
		&	Auc	&	0.793	&	0.855	&	0.843	&	0.828	&	0.856	&	\textbf{0.870}	\\
		&	Cvg	&	5.746	&	4.122	&	4.527	&	4.967	&	4.216	&	\textbf{3.834}	\\
		&	Ap	&	0.556	&	0.577	&	0.601	&	0.6	&	0.628	&	\textbf{0.635}	\\ \hline
		\multirow{4}{*}{Enr}&	Rkl	&	0.2	&	0.156	&	0.127	&	0.182	&	0.133	&	\textbf{0.123}	\\
		&	Auc	&	0.802	&	0.846	&	0.875	&	0.82	&	0.869	&	\textbf{0.877}	\\
		&	Cvg	&	23.336	&	19.006	&	\textbf{16.545}	&	21.974	&	17.82	&	17.2	\\
		&	Ap	&	0.56	&	0.557	&	0.622	&	0.55	&	0.578	&	\textbf{0.631}	\\ \hline
		\multirow{4}{*}{Ima}&	Rkl	&	0.213	&	0.185	&	0.184	&	0.186	&	0.222	&	\textbf{0.141}	\\
		&	Auc	&	0.787	&	0.815	&	0.816	&	0.814	&	0.778	&	\textbf{0.859}	\\
		&	Cvg	&	1.107	&	1.006	&	1.005	&	1.013	&	1.154	&	\textbf{0.831}	\\
		&	Ap	&	0.748	&	0.778	&	0.779	&	0.778	&	0.72	&	\textbf{0.825}	\\ \hline
		\multirow{4}{*}{Soc}&	Rkl	&	0.13	&	0.084	&	0.079	&	0.081	&	0.064	&	\textbf{0.052}	\\
		&	Auc	&	0.87	&	0.916	&	0.921	&	0.919	&	0.936	&	\textbf{0.948}	\\
		&	Cvg	&	6.661	&	4.393	&	4.088	&	4.500	&	3.634	&	\textbf{2.975}	\\
		&	Ap	&	0.697	&	0.595	&	0.671	&	0.757	&	0.776	&	\textbf{0.780}	\\ \hline
		\multirow{4}{*}{Cor}&	Rkl	&	0.166	&	0.155	&	0.181	&	0.146	&	0.15	&	\textbf{0.143}	\\
		&	Auc	&	0.834	&	0.845	&	0.819	&	\textbf{0.858}	&	0.850	&	0.857	\\
		&	Cvg	&	48.697	&	46.478	&	51.574	&	43.637	&	44.996	&	\textbf{43.137}	\\
		&	Ap	&	0.326	&	0.284	&	0.300	&	0.334	&	0.323	&	\textbf{0.346}	\\  \hline
		\multirow{4}{*}{Tmc}&	Rkl	&	0.047	&	0.045	&	0.047	&	0.047	&	0.046	&	\textbf{0.031}	\\
		&	Auc	&	0.953	&	0.955 &	0.953	&	0.953	&	0.955	&	\textbf{0.969}	\\
		&	Cvg	&	2.922	&	2.811	&	2.906	&	2.928	&	2.824	&	\textbf{2.337}	\\
		&	Ap	&	0.832	&	0.831	&	0.83	&	0.832	&	0.83	&	\textbf{0.881}	\\
		\hline  \hline
	\end{tabular}
\end{table*}

\begin{table*}[htp!]
	\caption{Results for learning with missing labels ($\rho = 0.3$).}
	\label{miss3_result}
	\centering
	\scriptsize
	\begin{tabular}{ccccccccc}
		\hline \hline
		&  & LEML   & ML-LRC & GLOCAL & LSML   & DM2L-l & DM2L-nl\\
		\hline
		\multirow{4}{*}{Art}&	Rkl	&	0.234	&	0.167	&	0.178	&	0.167	&	0.143	&	\textbf{0.127}	\\
		&	Auc	&	0.768	&	0.835	&	0.825	&	0.836	&	0.859	&	\textbf{0.873}	\\
		&	Cvg	&	8.237	&	5.849	&	6.62	&	6.329	&	5.618	&	\textbf{5.125}	\\
		&	Ap	&	0.505	&	0.468	&	0.568	&	0.585	&	0.61	&	\textbf{0.614}	\\ \hline
		\multirow{4}{*}{Bus}&	Rkl	&	0.101	&	0.061	&	0.043	&	0.073	&	0.046	&	\textbf{0.040}	\\
		&	Auc	&	0.901	&	0.94	&	0.958	&	0.928	&	0.955	&	\textbf{0.960}	\\
		&	Cvg	&	5.215	&	3.46	&	\textbf{2.426}	&	3.999	&	2.775	&	2.546	\\
		&	Ap	&	0.777	&	0.856	&	0.872	&	0.835	&	0.876	&	\textbf{0.886}	\\ \hline
		\multirow{4}{*}{Rec}&	Rkl	&	0.2402	&	0.1674	&	0.181	&	0.203	&	0.152	&	\textbf{0.137}	\\
		&	Auc	&	0.7638	&	0.837	&	0.823	&	0.801	&	0.852	&	\textbf{0.863}	\\
		&	Cvg	&	6.4295	&	4.5927	&	4.879	&	5.608	&	4.378	&	\textbf{4.000}	\\
		&	Ap	&	0.5164	&	0.5646	&	0.527	&	0.569	&	\textbf{0.617}	&	0.61	\\ \hline
		\multirow{4}{*}{Enr}&	Rkl	&	0.259	&	0.212	&	0.156	&	0.26	&	\textbf{0.144}	&	0.147	\\
		&	Auc	&	0.743	&	0.791	&	0.846	&	0.742	&	\textbf{0.858}	&	0.853	\\
		&	Cvg	&	28.42	&	23.666	&	19.533	&	27.283	&	\textbf{18.853}	&	19.691	\\
		&	Ap	&	0.487	&	0.471	&	0.576	&	0.406	&	0.553	&	\textbf{0.583}	\\ \hline
		\multirow{4}{*}{Ima}&	Rkl	&	0.231	&	0.196	&	0.2	&	0.201	&	0.243	&	\textbf{0.150}	\\
		&	Auc	&	0.769	&	0.805	&	0.8	&	0.799	&	0.757	&	\textbf{0.850}	\\
		&	Cvg	&	1.184	&	1.054	&	1.071	&	1.078	&	1.245	&	\textbf{0.872}	\\
		&	Ap	&	0.729	&	0.768	&	0.759	&	0.764	&	0.697	&	\textbf{0.816}	\\ \hline
		\multirow{4}{*}{Soc}&	Rkl	&	0.153	&	0.088	&	0.081	&	0.098	&	0.066	&	\textbf{0.055}	\\
		&	Auc	&	0.847	&	0.912	&	0.919	&	0.902	&	0.934	&	\textbf{0.945}	\\
		&	Cvg	&	7.67	&	4.593	&	4.206	&	5.278	&	3.819	&	\textbf{3.172}	\\
		&	Ap	&	0.671	&	0.594	&	0.671	&	0.735	&	\textbf{0.768}	&	\textbf{0.768}	\\ \hline
		\multirow{4}{*}{Cor}&	Rkl	&	0.196	&	0.172	&	0.208	&	0.203	&	0.179	&	\textbf{0.155}	\\
		&	Auc	&	0.804	&	0.829	&	0.793	&	0.834	&	0.821	&	\textbf{0.845}	\\
		&	Cvg	&	57.347	&	52.244	&	58.363	&	51.9	&	52.7	&	\textbf{47.274}	\\
		&	Ap	&	0.299	&	0.286	&	0.282	&	\textbf{0.318}	&	0.282	&	0.311	\\ \hline
		\multirow{4}{*}{Tmc}&	Rkl	&	0.057	&	0.05	&	0.053	&	0.058	&	0.049	&	\textbf{0.039}	\\
		&	Auc	&	0.943	&	0.95	&	0.948	&	0.942	&	0.951	&	\textbf{0.961}	\\
		&	Cvg	&	3.336	&	2.944	&	3.126	&	3.384	&	2.957	&	\textbf{2.666}	\\
		&	Ap	&	0.818	&	0.805	&	0.821	&	0.816	&	0.823	&	\textbf{0.864}	\\
		\hline  \hline
	\end{tabular}
\end{table*}

\begin{figure}[htbp]
	\centering
	\subfigure[Ranking Loss]{
		\centering
		\includegraphics[height=4.5cm,width=7cm]{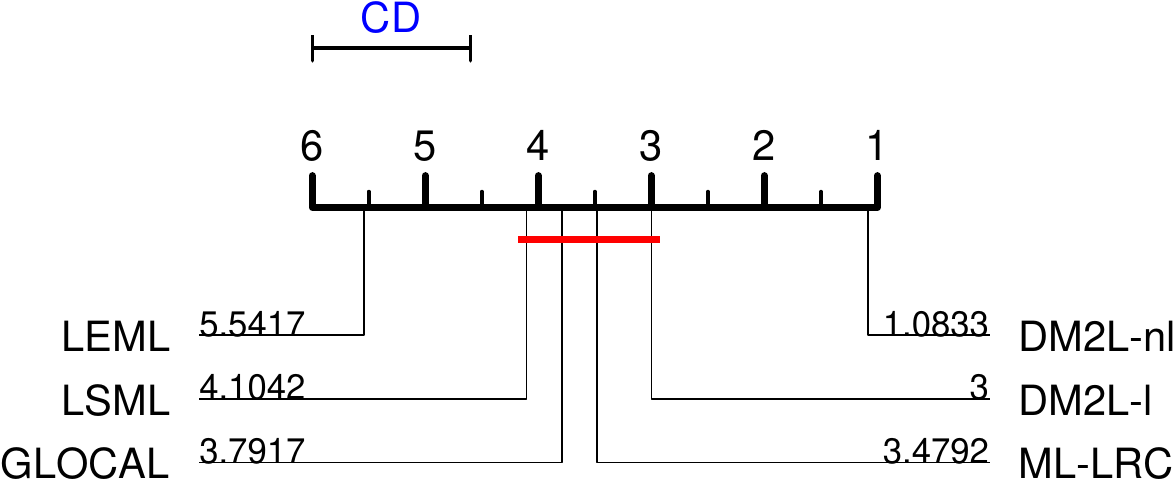}
	}%
	\subfigure[AUC]{
		\includegraphics[height=4.5cm,width=7cm]{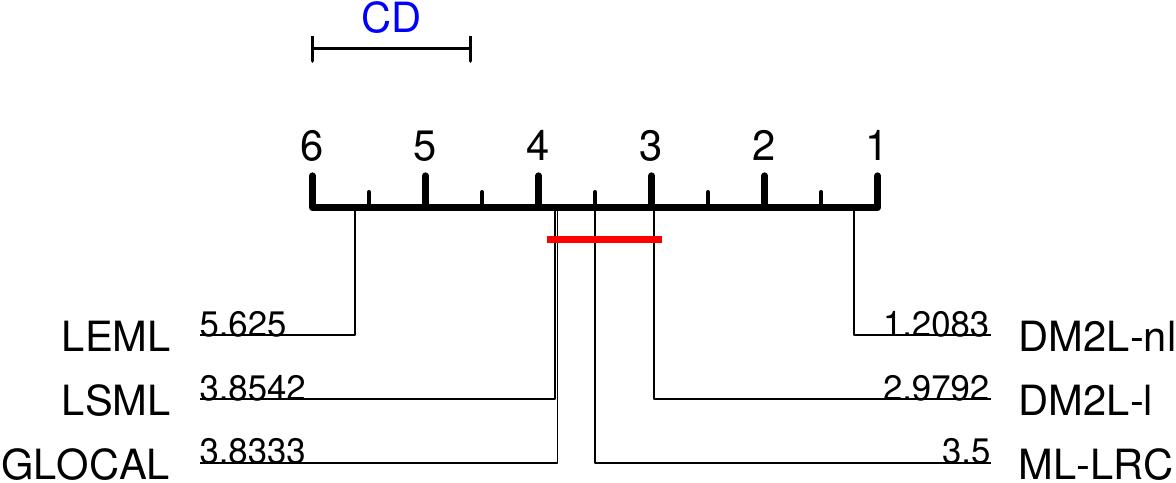}
	}%
	
	\subfigure[Coverage]{
		\includegraphics[height=4.5cm, width=7cm]{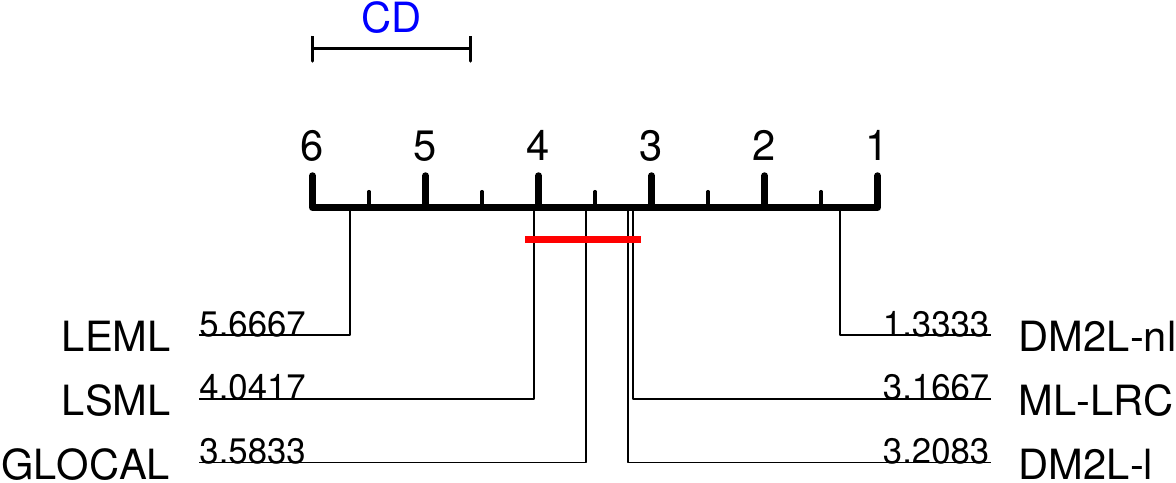}
	}%
	\subfigure[Average Precison]{
		\includegraphics[height=4.5cm,width=7cm]{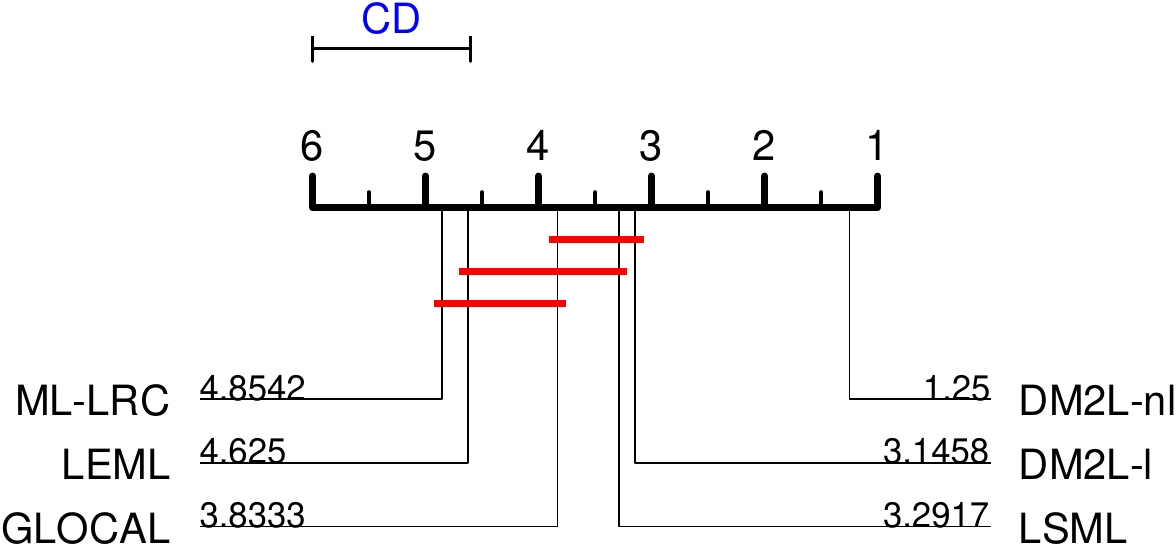}
	}%
	\caption{Comparison of DM2L-nl and DM2L-l against other comparing algorithms with the Nemenyi test. Groups of classifiers that are not significantly different (at $\alpha = 0.05$) are connected. }
	\label{nemenyitest}
\end{figure}

{{To further analyze their relative performance, the Nemenyi test is employed. As the missing rate is varied from $\{0, 30\%, 70\%\}$, there are 24 (8 datasets $\times$ 3 missing rates) points totally. The performances between two classifiers will be significantly different if their average ranks differ by at least one critical difference $CD = q_{\alpha} \sqrt{k(k+1)/6N}$. For Nemenyi test, $q_{\alpha} = 3.102$ at significance level $\alpha = 0.05$, and thus $CD= 1.6753 (k=6, N=24)$. Fig. \ref{nemenyitest} shows the CD diagrams on each evaluation metric. In each sub-figure, any compared algorithms whose average ranks differ within one CD are connected. Otherwise, any pair of algorithms not connected is considered to significantly differ in performance.}}

{{As can be seen from Fig. \ref{nemenyitest}, 1) DM2L-l is significantly different from LEML on all evaluation measures, but LSML and GLOCAL are not on Average Precison. Thus, even involving the fewest assumptions and only one hyper-parameter, DM2L-l achieves slightly better performance than ML-LRC, LEML, GLOCAL and LSML. This indicates the effectiveness of jointly exploring global label structure, local label structure and the discriminant information to a certain extent. 2) DM2L-nl, DM2L-l and GLOCAL perform better than LEML with significant difference on Ranking Loss, AUC and Coverage, indicating the importance of exploiting local label structure. 3) DM2L-l is better than GLOCAL on all evaluation measures, indicating the better effectiveness of local low-rank label structure than hypothetical local label structure. 4) DM2L-nl performs best on all evaluation measures and also shows significant differences from other methods, indicating the powerful nonlinear representation ability of DM2L-nl. In the following, we will further analyze the importances of exploiting local low-rank label structure and label discriminant information.}}

\subsection{Analysis}
\subsubsection{Effectiveness of local label structure}
To show the effectiveness of local label structures, the following two models are compared,
\begin{itemize}
	\item DM2L-Lo(DM2L model only involving local label structure term), namely,
	\begin{equation}
	\min_{\mathbf{W}} \frac{1}{2}\|\mathcal{R}_\Omega(\mathbf{XW}) - \tilde{\mathbf{Y}}\|_F^2 + \lambda_d \sum_{k=1}^c \|\mathbf{X_kW}\|_*
	\end{equation}
	\item LEML (Low-rank empirical risk minimization for multi-label learning)\cite{yu2014large}. It learns a linear instance-to-label mapping with low-rank structure to take advantage of global label structure.
\end{itemize}

\begin{figure}[htbp!]
	\subfigure[Rkl $\downarrow$.]{
		\centering
		\includegraphics[height=3.7cm,width=4.5cm]{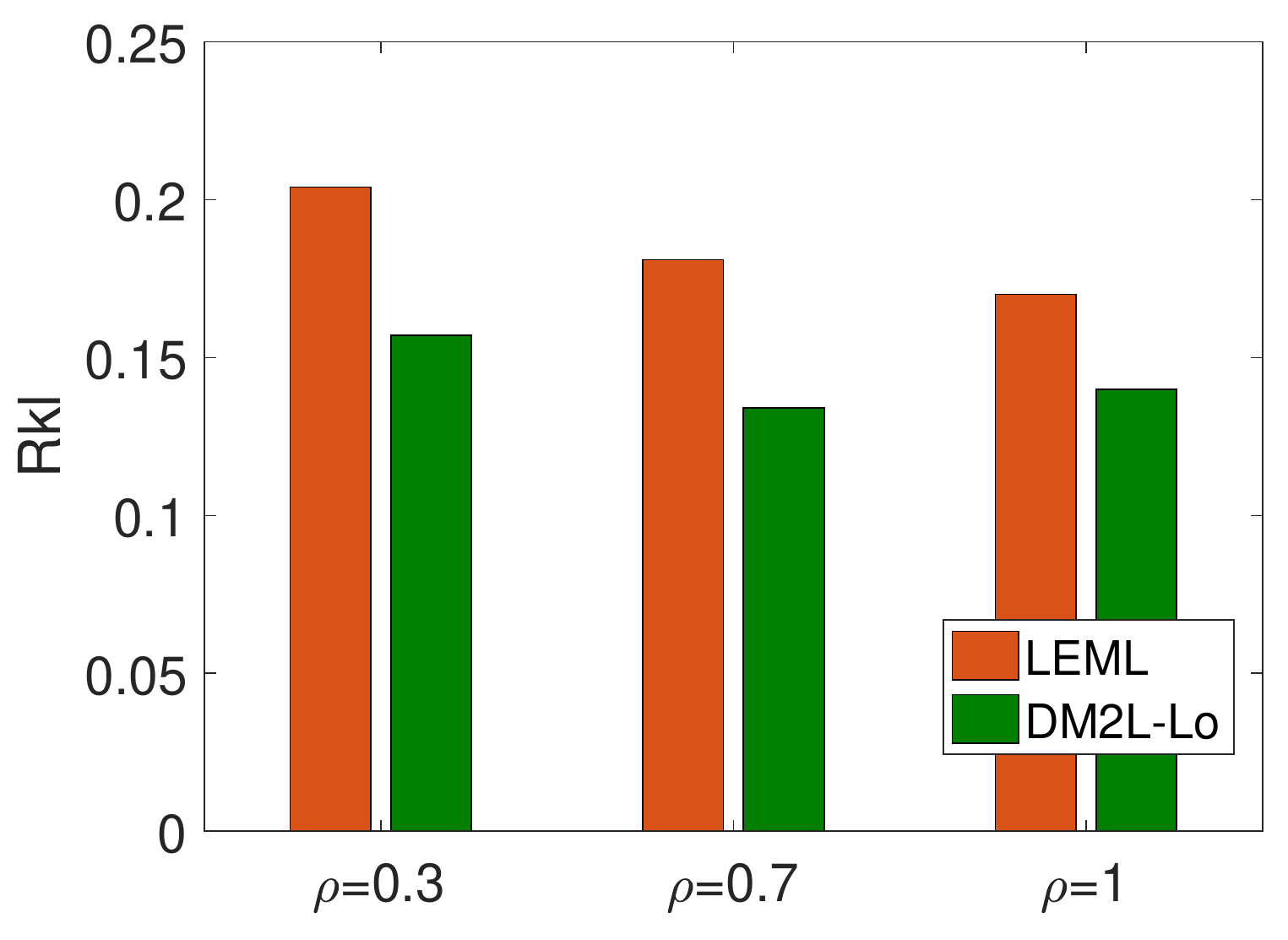}
	}%
	\subfigure[AUC $\uparrow$.]{
		\includegraphics[height=3.7cm,width=4.5cm]{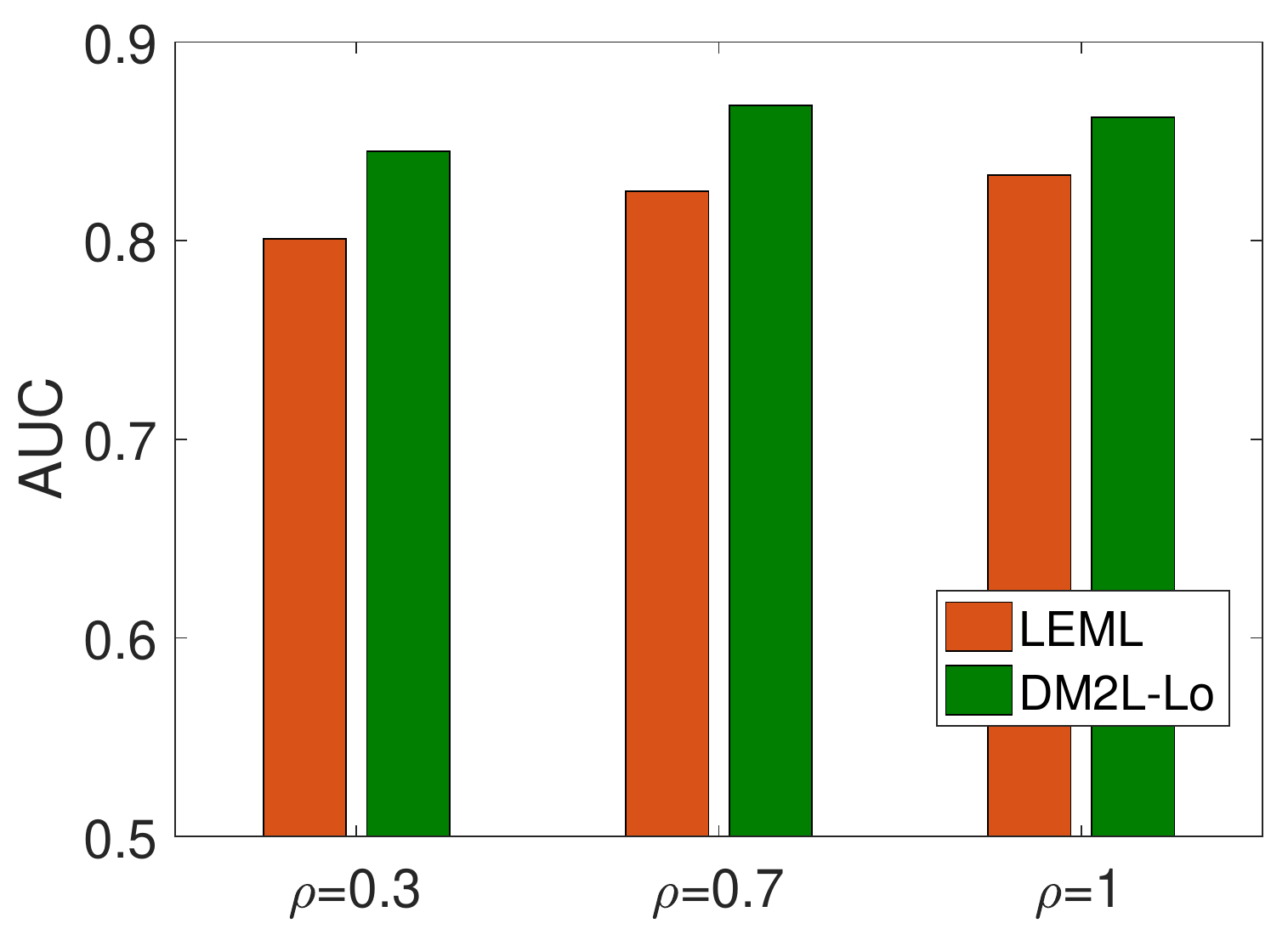}
	}%
	
	\subfigure[Cvg $\downarrow$.]{
		\includegraphics[height=3.7cm, width=4.5cm]{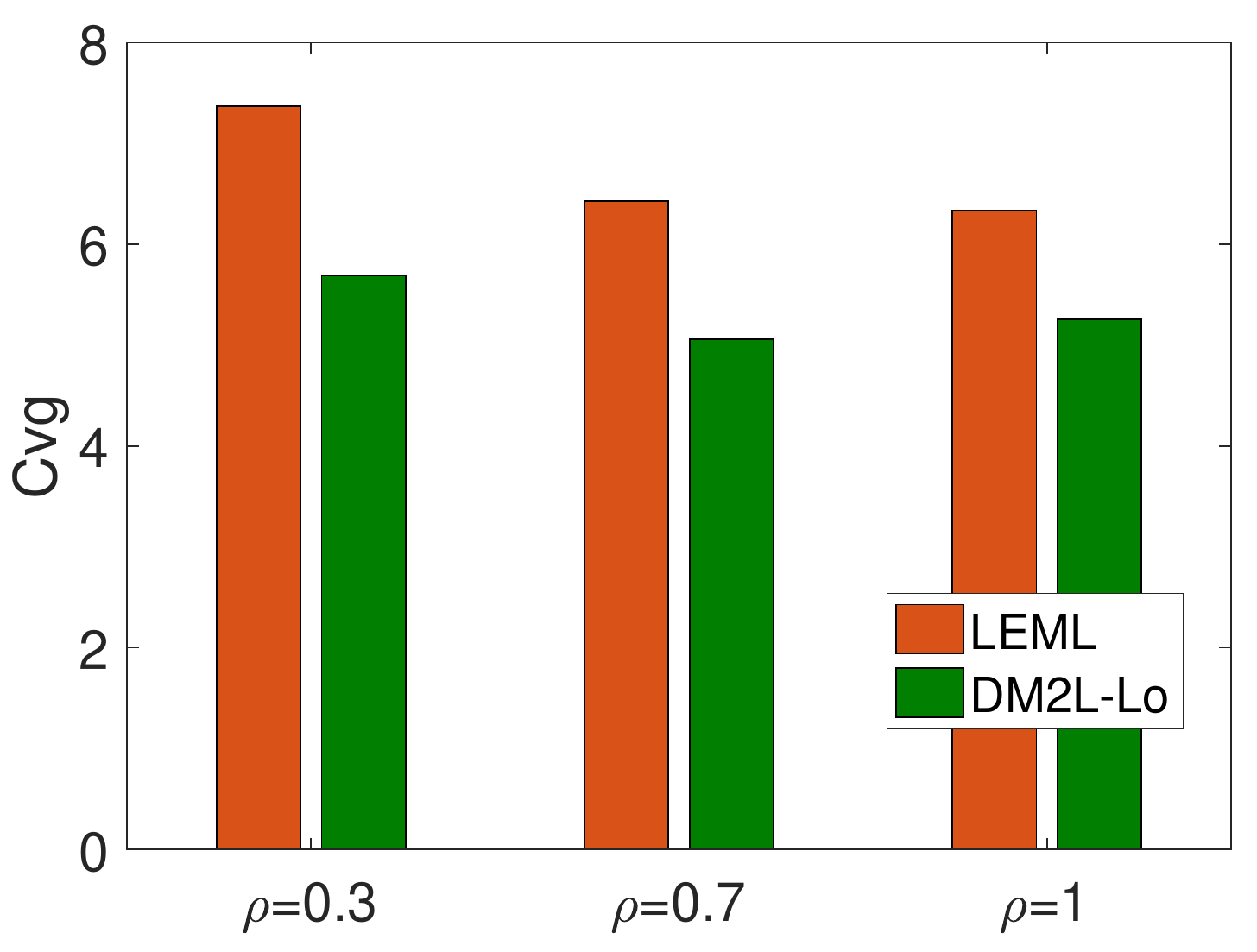}
	}%
	\subfigure[Ap $\uparrow$.]{
		\includegraphics[height=3.7cm,width=4.5cm]{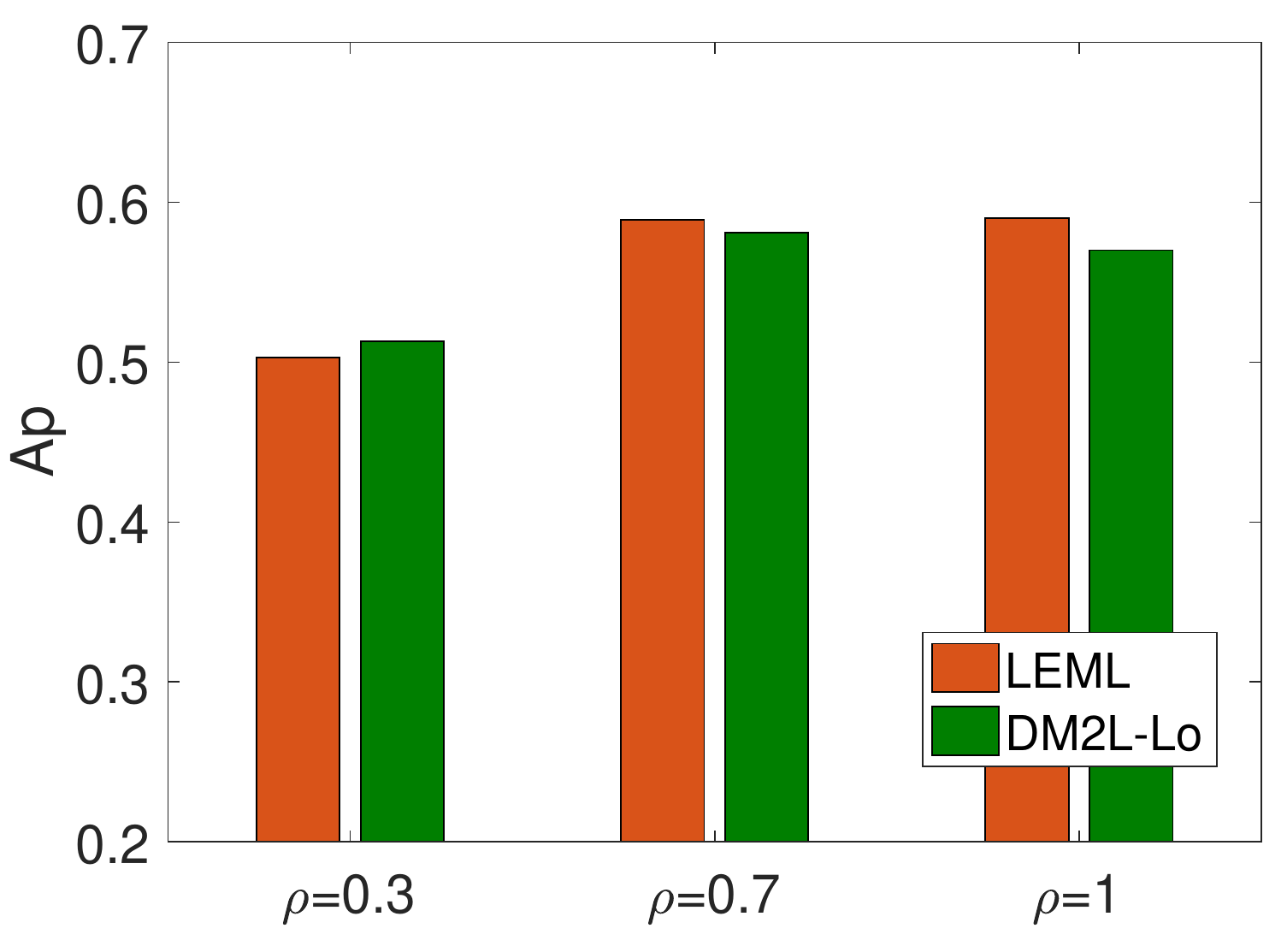}
	}%
	\centering
	\caption{LEML vs. DM2L-Lo on Arts dataset.}
	\label{local_ana1}
\end{figure}

\begin{figure}[htbp!]
	\subfigure[Rkl $\downarrow$.]{
		\centering
		\includegraphics[height=3.7cm,width=4.5cm]{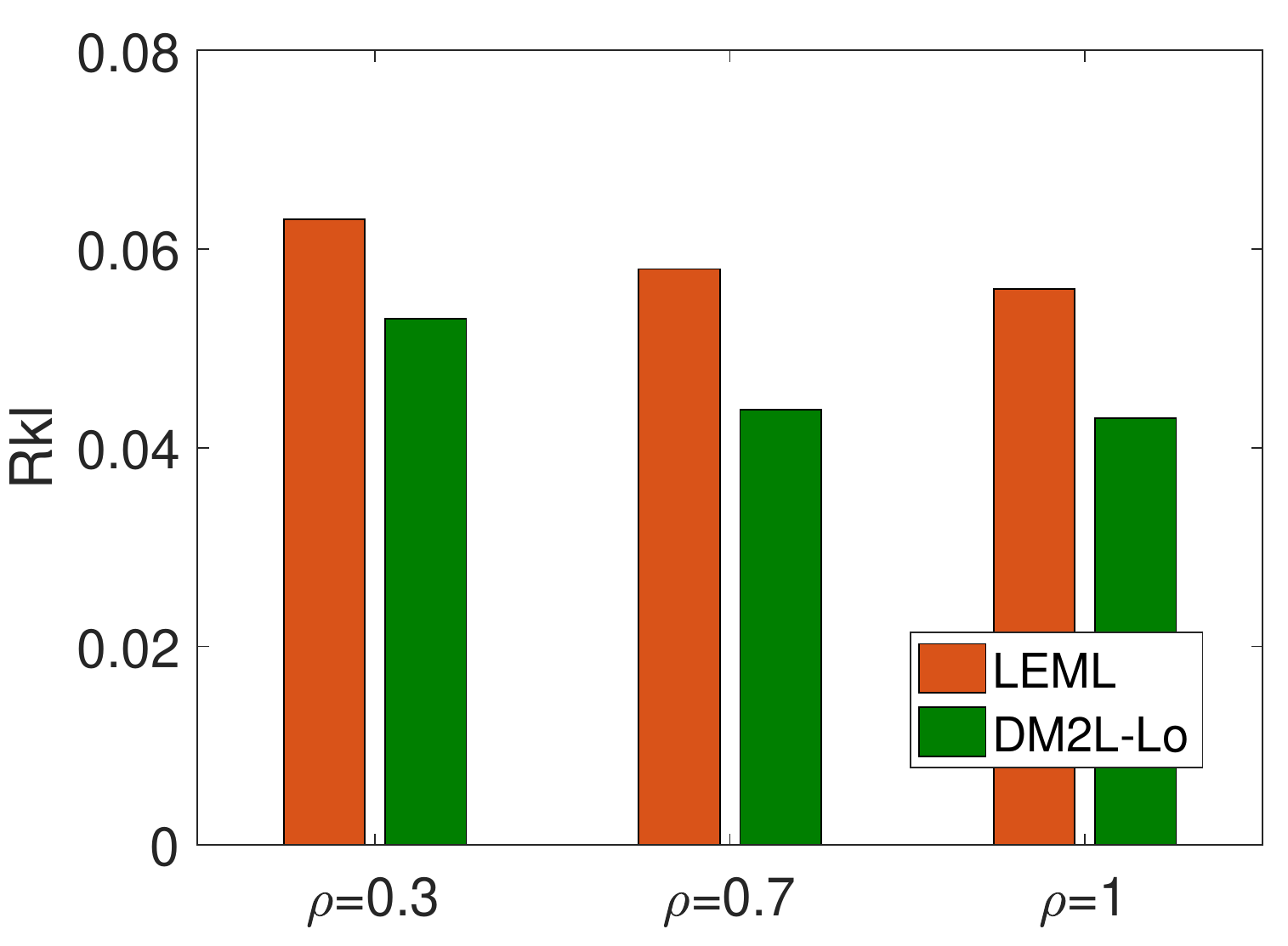}
	}%
	\subfigure[AUC $\uparrow$.]{
		\includegraphics[height=3.7cm,width=4.5cm]{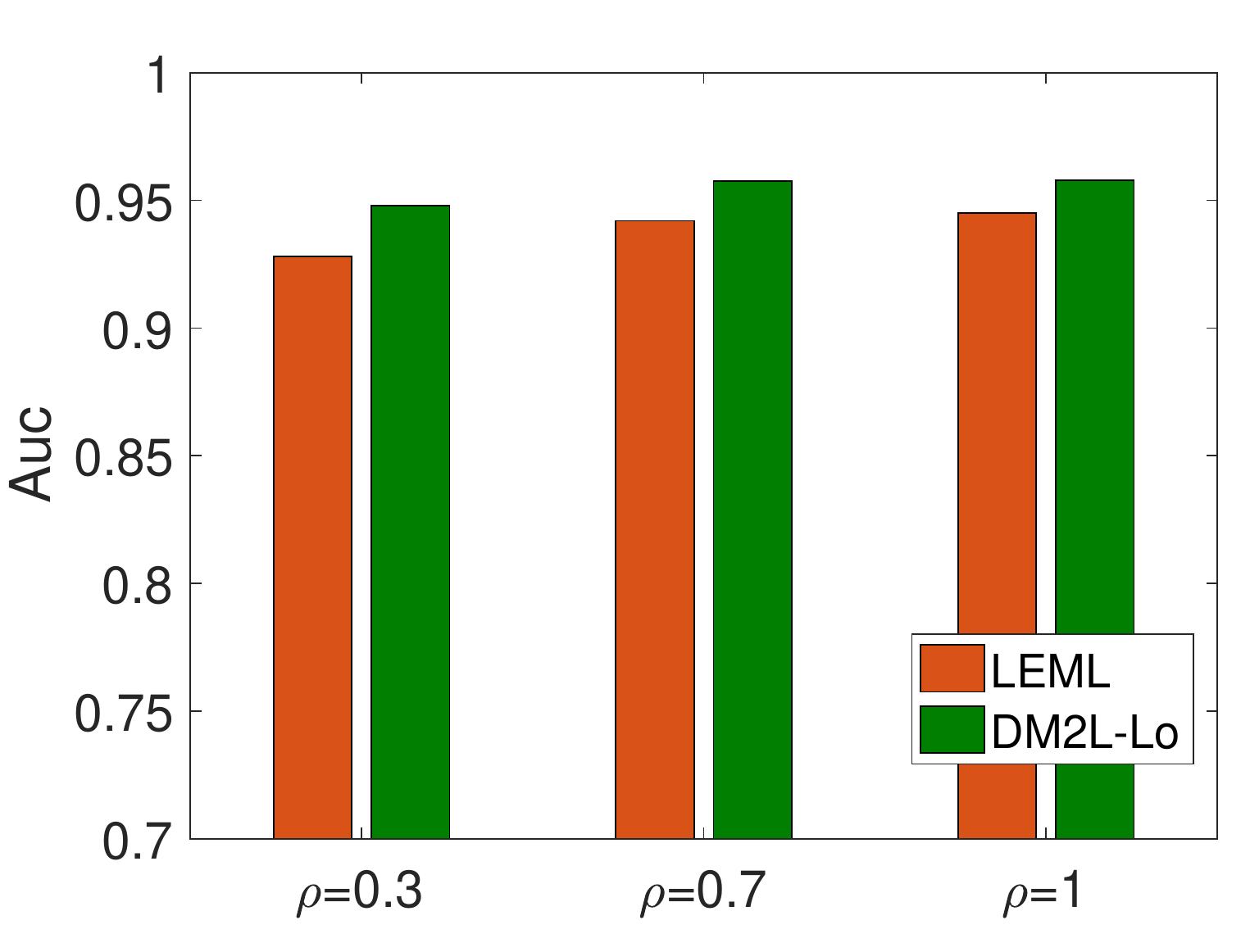}
	}%
	
	\subfigure[Cvg $\downarrow$.]{
		\includegraphics[height=3.7cm, width=4.5cm]{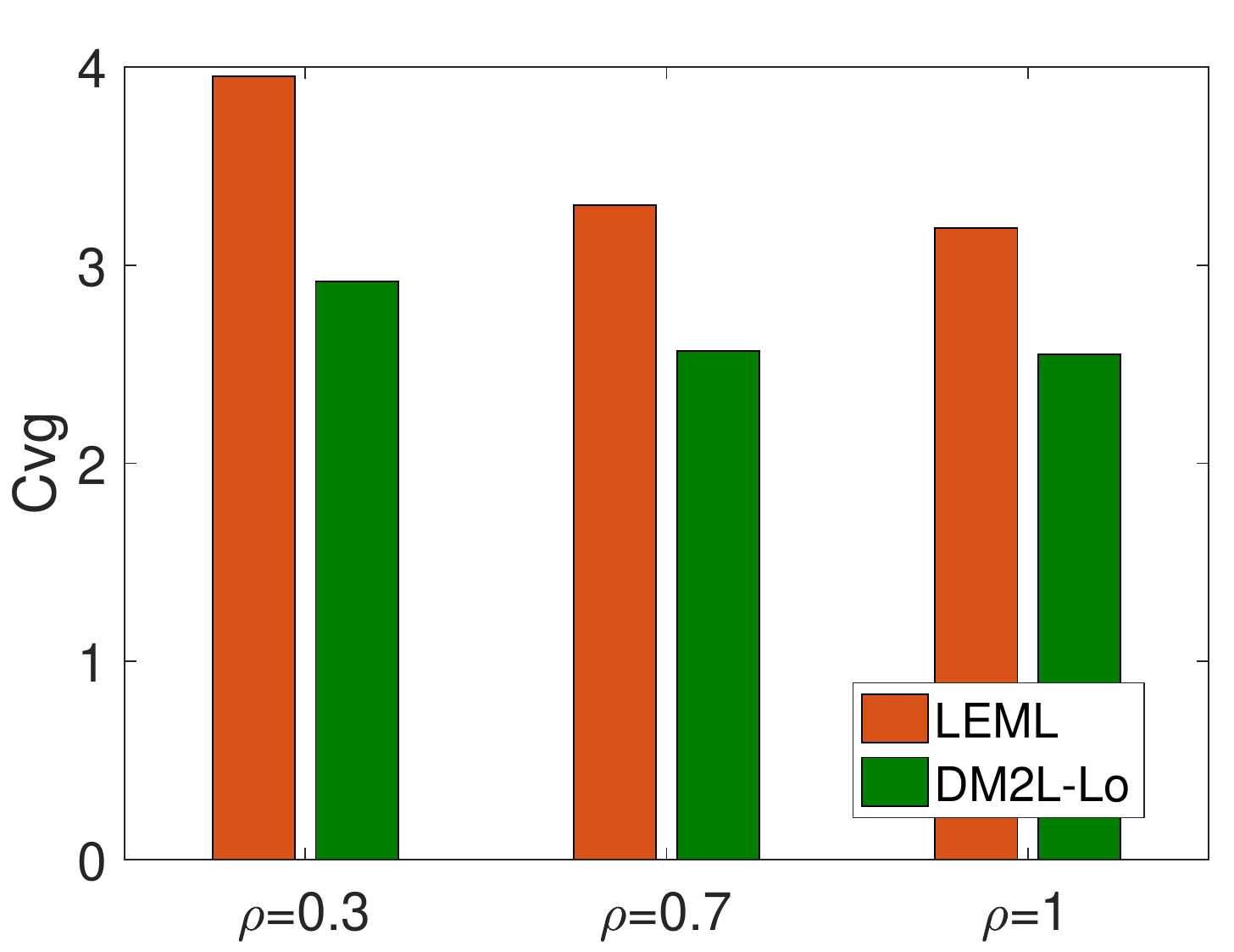}
	}%
	\subfigure[Ap $\uparrow$.]{
		\includegraphics[height=3.7cm,width=4.5cm]{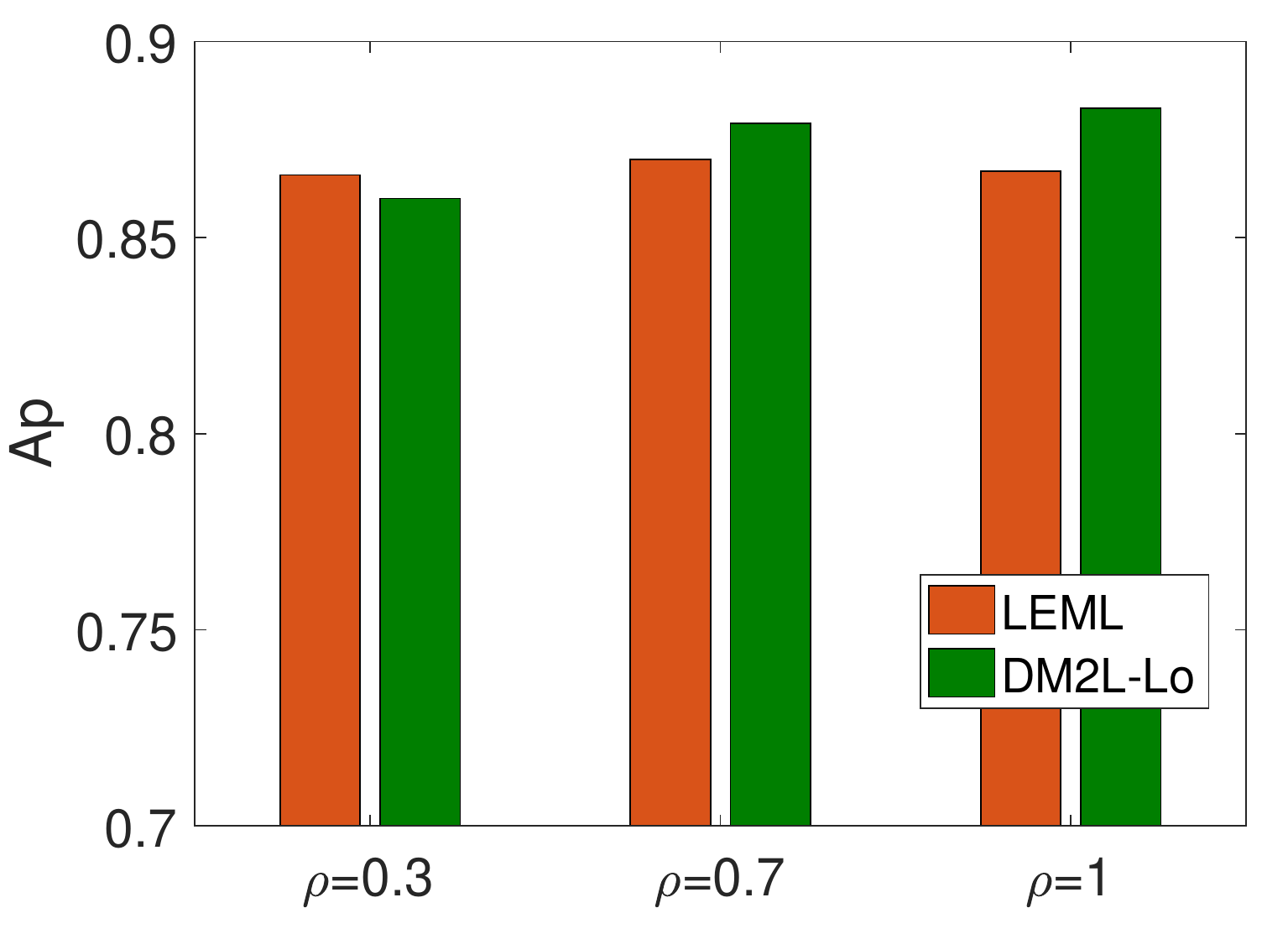}
	}%
	\centering
	\caption{LEML vs. DM2L-Lo on Business dataset.}
	\label{local_ana2}
\end{figure}

The experiments are conducted on the Arts and Business datasets with both full and missing labels. Fig. \ref{local_ana1} and Fig. \ref{local_ana2}  show the results in terms of four evaluation measures. As can be seen, DM2L-Lo almost beats LEML in terms of all the evaluation measures and a similar phenomenon on the other datasets can be also seen. This reflects the importance of local label structure both for multi-label classification with full and missing labels.

\subsubsection{Effectiveness of discriminant label information}
\begin{figure}[htbp!]
	\subfigure[Rkl $\downarrow$.]{
		\centering
		\includegraphics[height=3.7cm,width=4.5cm]{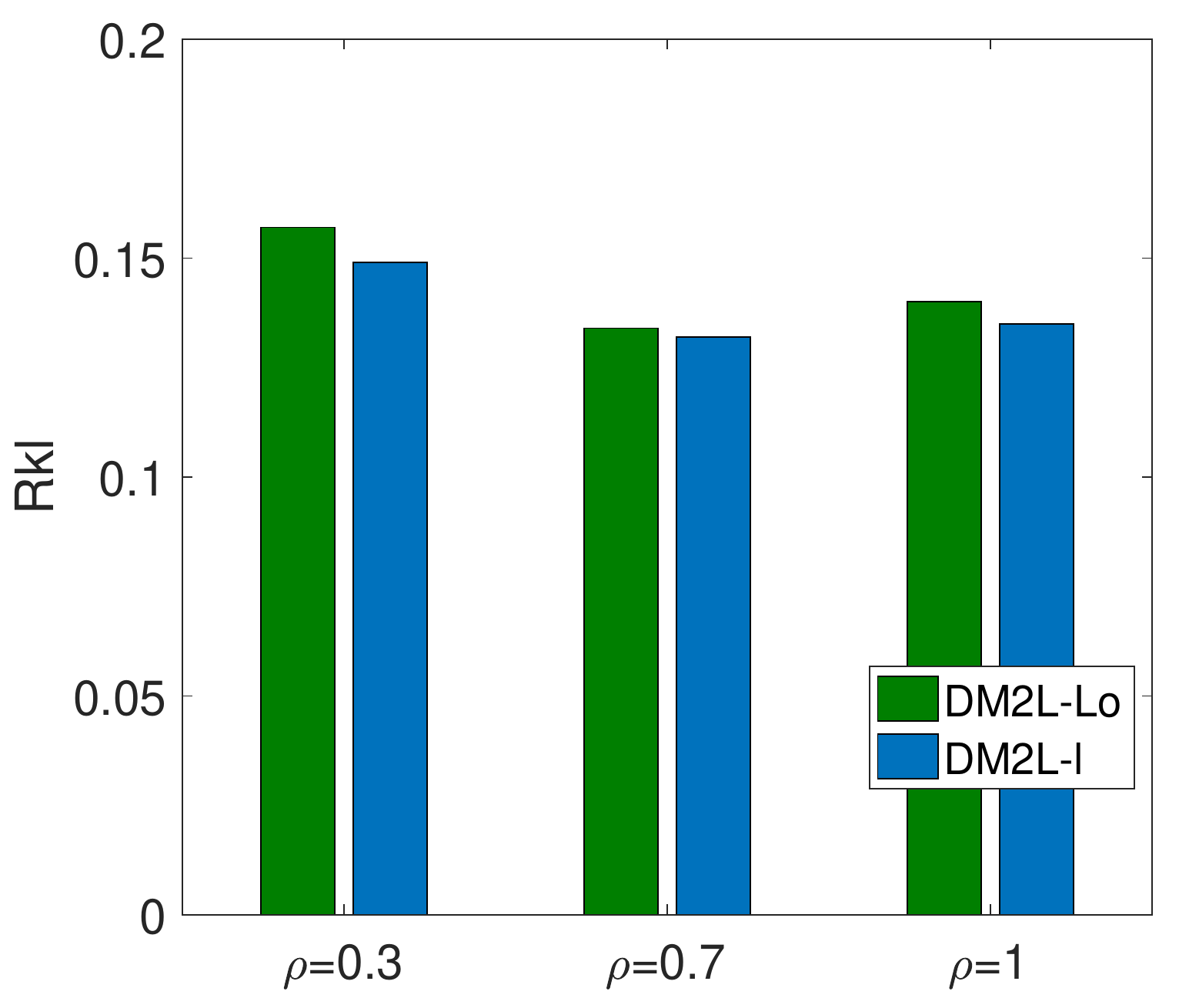}
	}%
	\subfigure[AUC $\uparrow$.]{
		\includegraphics[height=3.7cm,width=4.5cm]{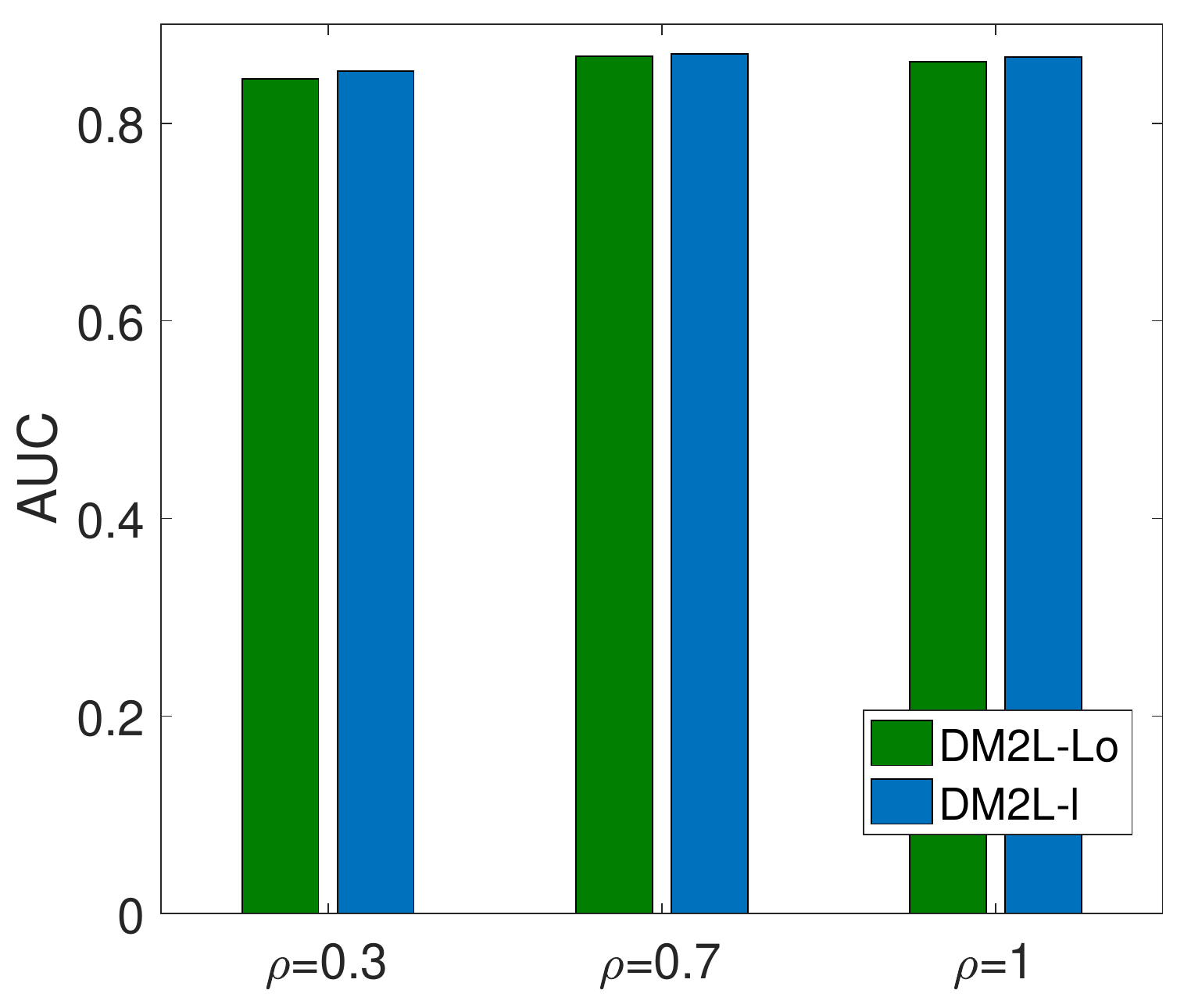}
	}%
	
	\subfigure[Cvg $\downarrow$.]{
		\includegraphics[height=3.7cm, width=4.5cm]{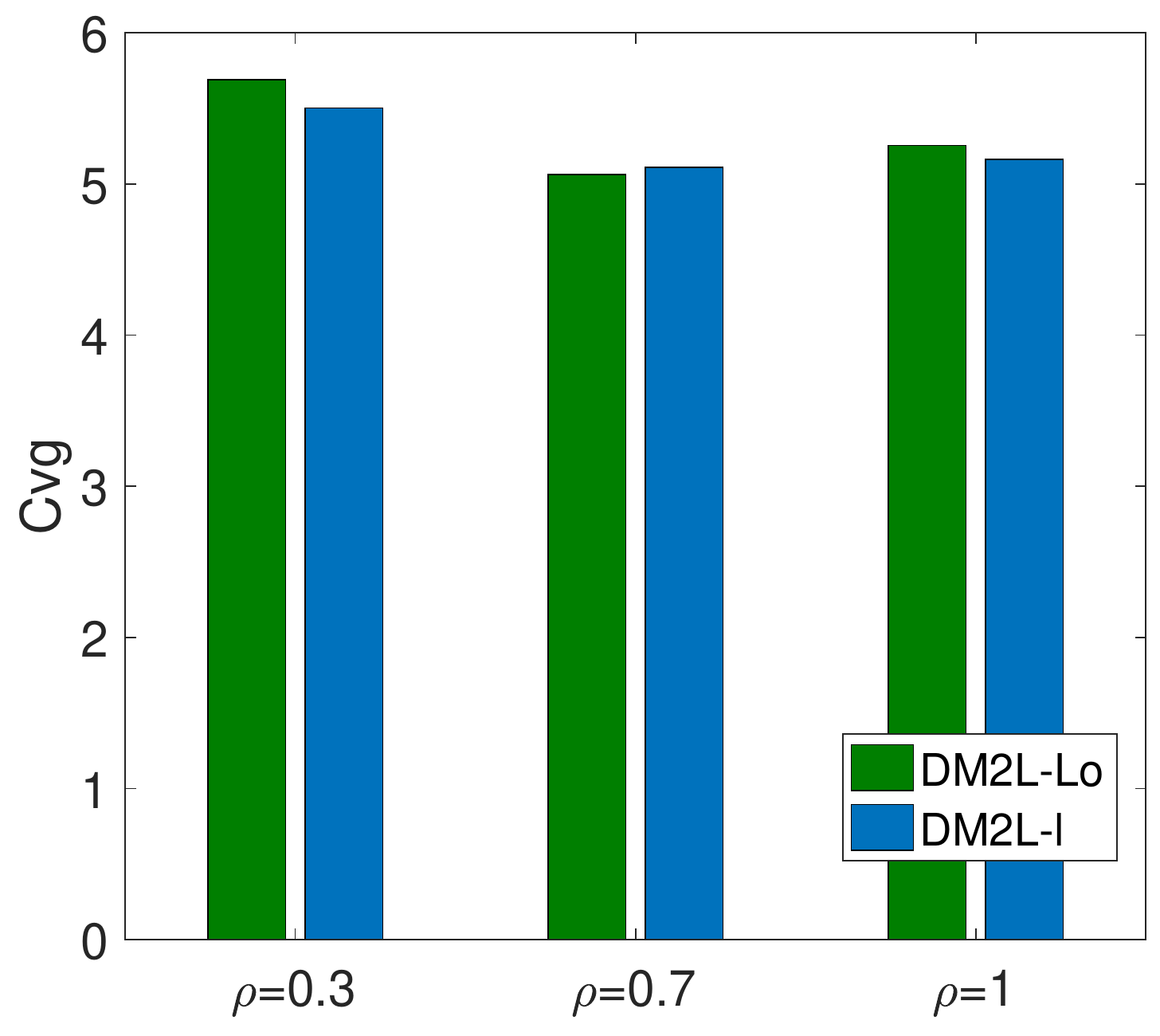}
	}%
	\subfigure[Ap $\uparrow$.]{
		\includegraphics[height=3.7cm,width=4.5cm]{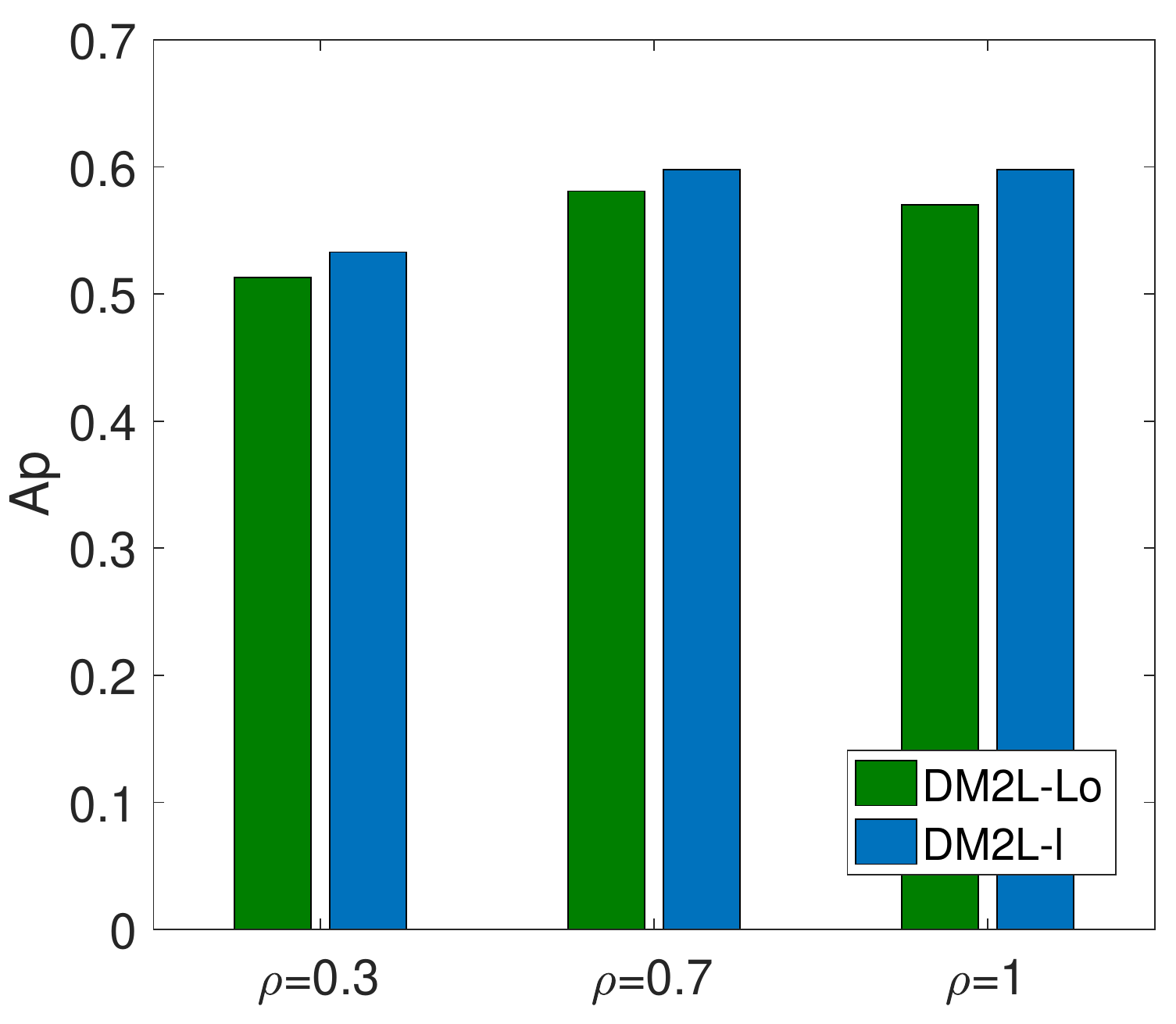}
	}%
	\centering
	\caption{DM2L-l vs. DM2L-Lo on Arts dataset.}
	\label{disResu1}
\end{figure}

\begin{figure}[htbp!]
	\subfigure[Rkl $\downarrow$.]{
		\centering
		\includegraphics[height=3.7cm,width=4.5cm]{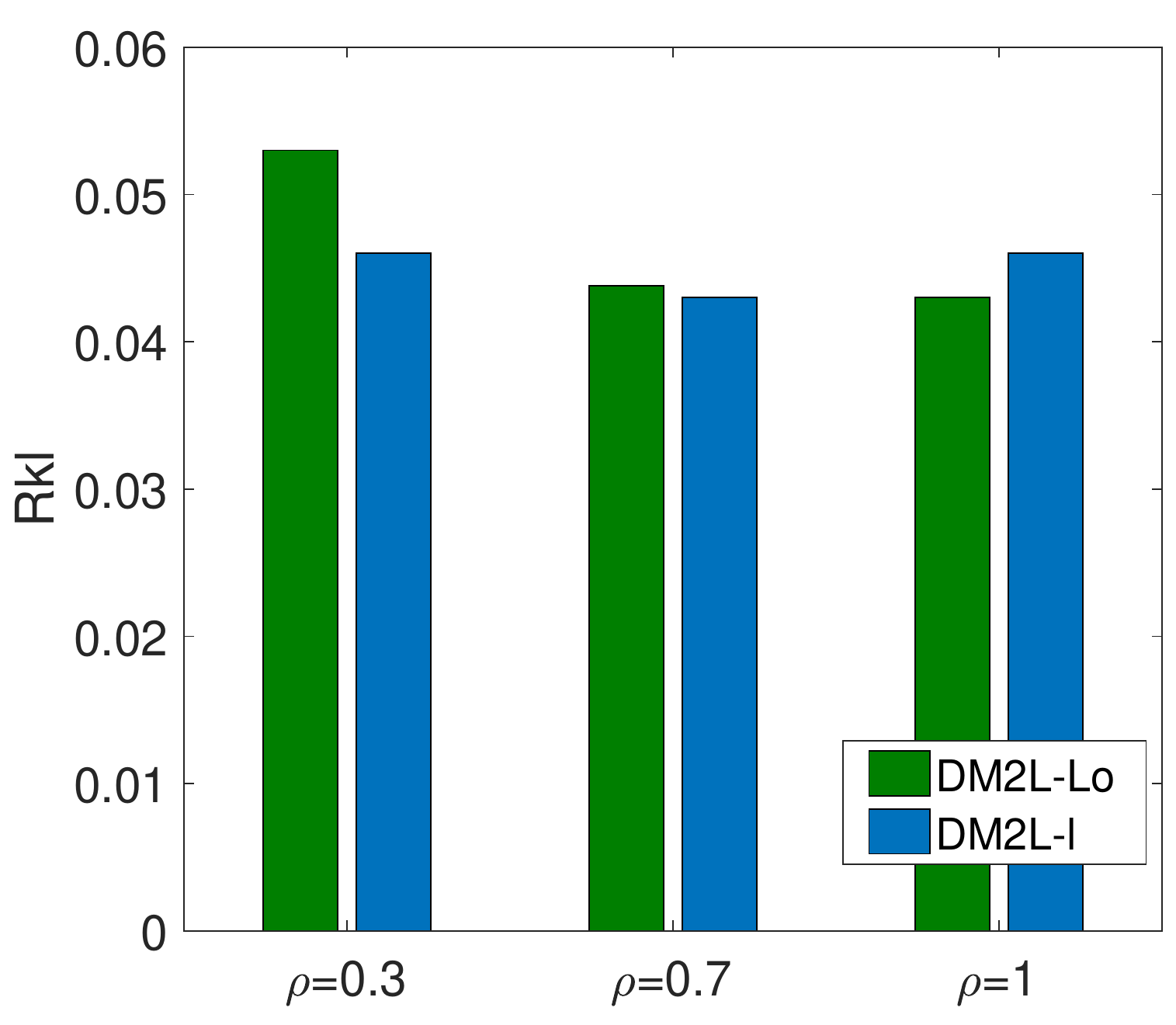}
	}%
	\subfigure[AUC $\uparrow$.]{
		\includegraphics[height=3.7cm,width=4.5cm]{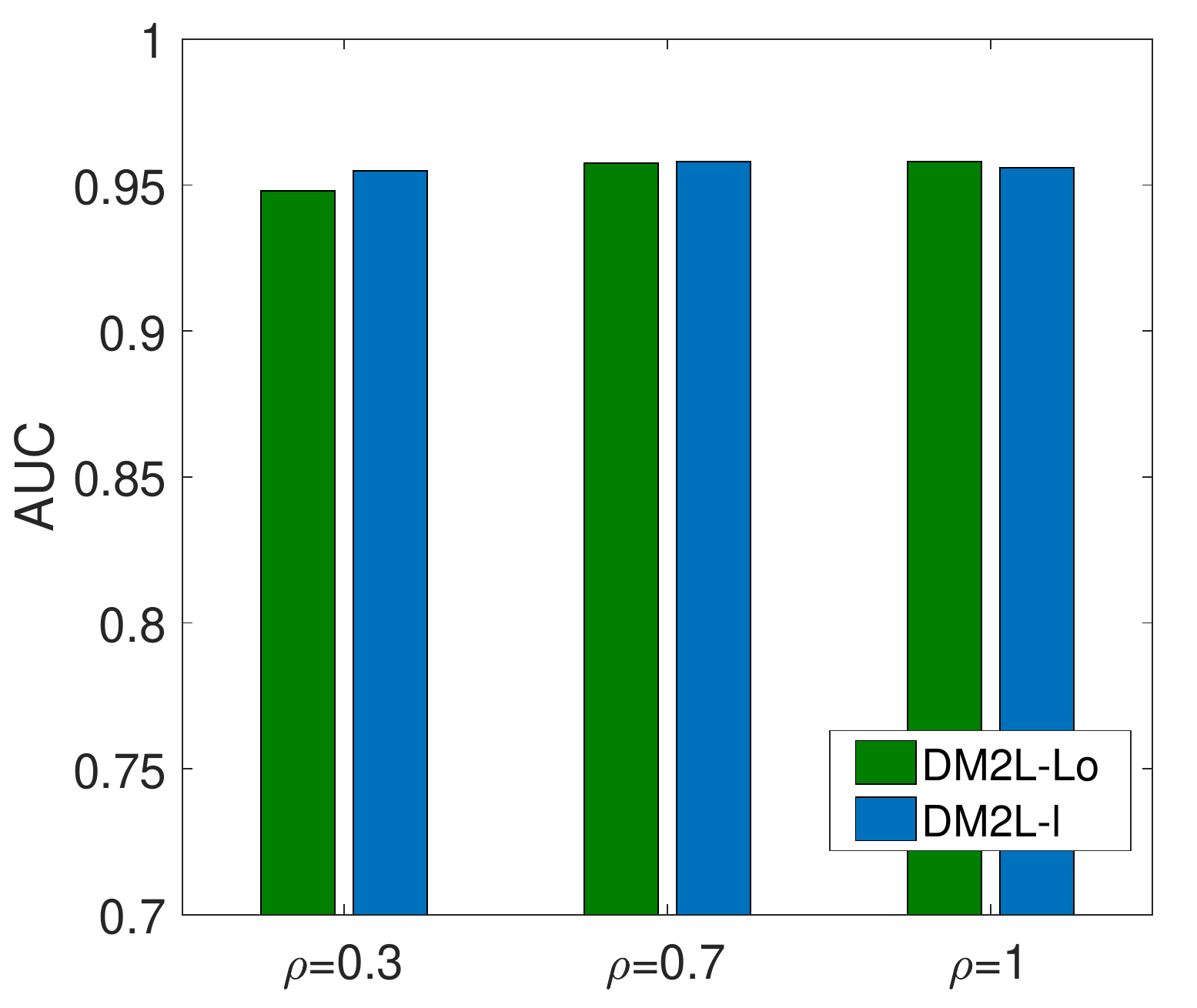}
	}%
	
	\subfigure[Cvg $\downarrow$.]{
		\includegraphics[height=3.7cm, width=4.5cm]{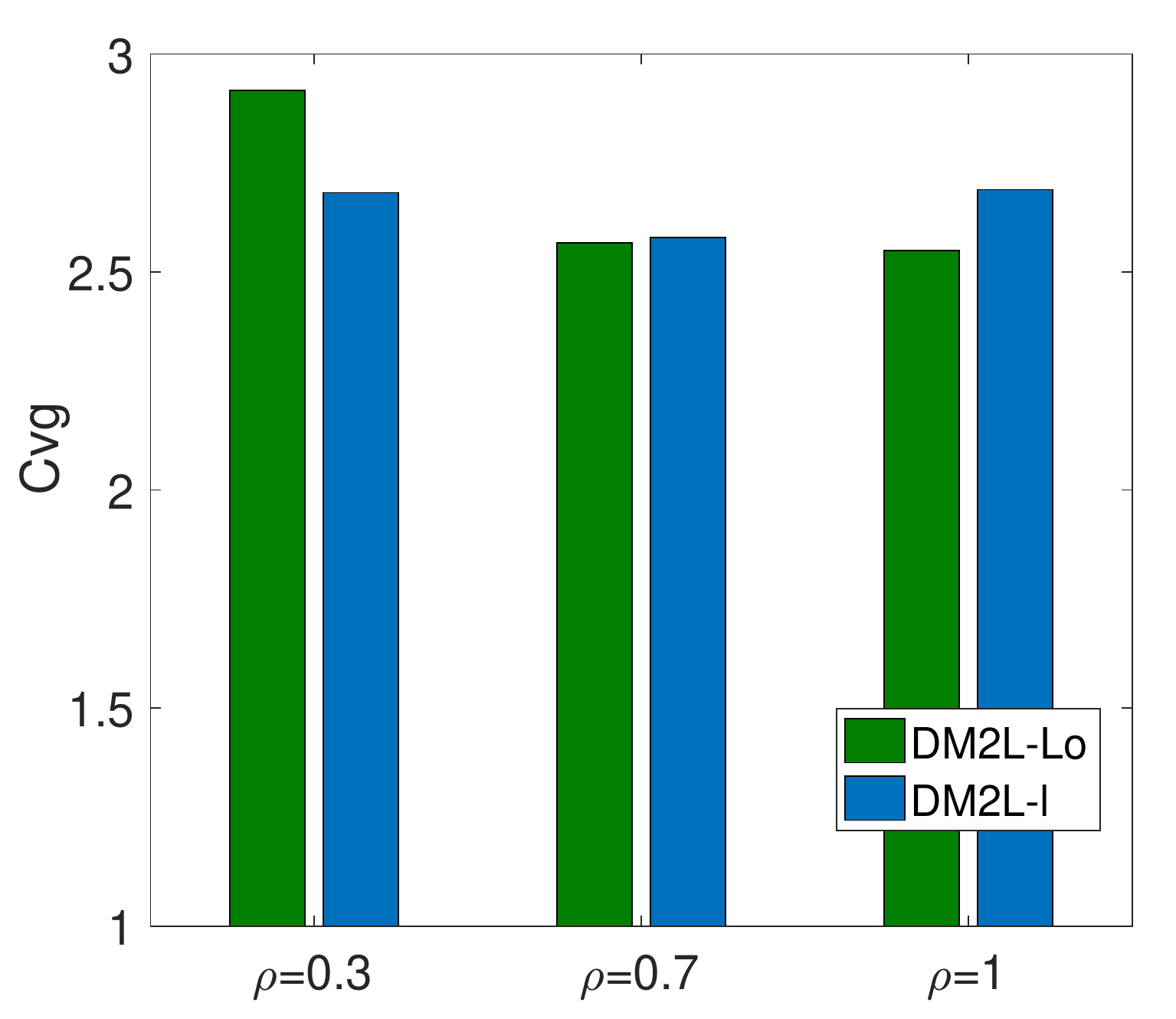}
	}%
	\subfigure[Ap $\uparrow$.]{
		\includegraphics[height=3.7cm,width=4.5cm]{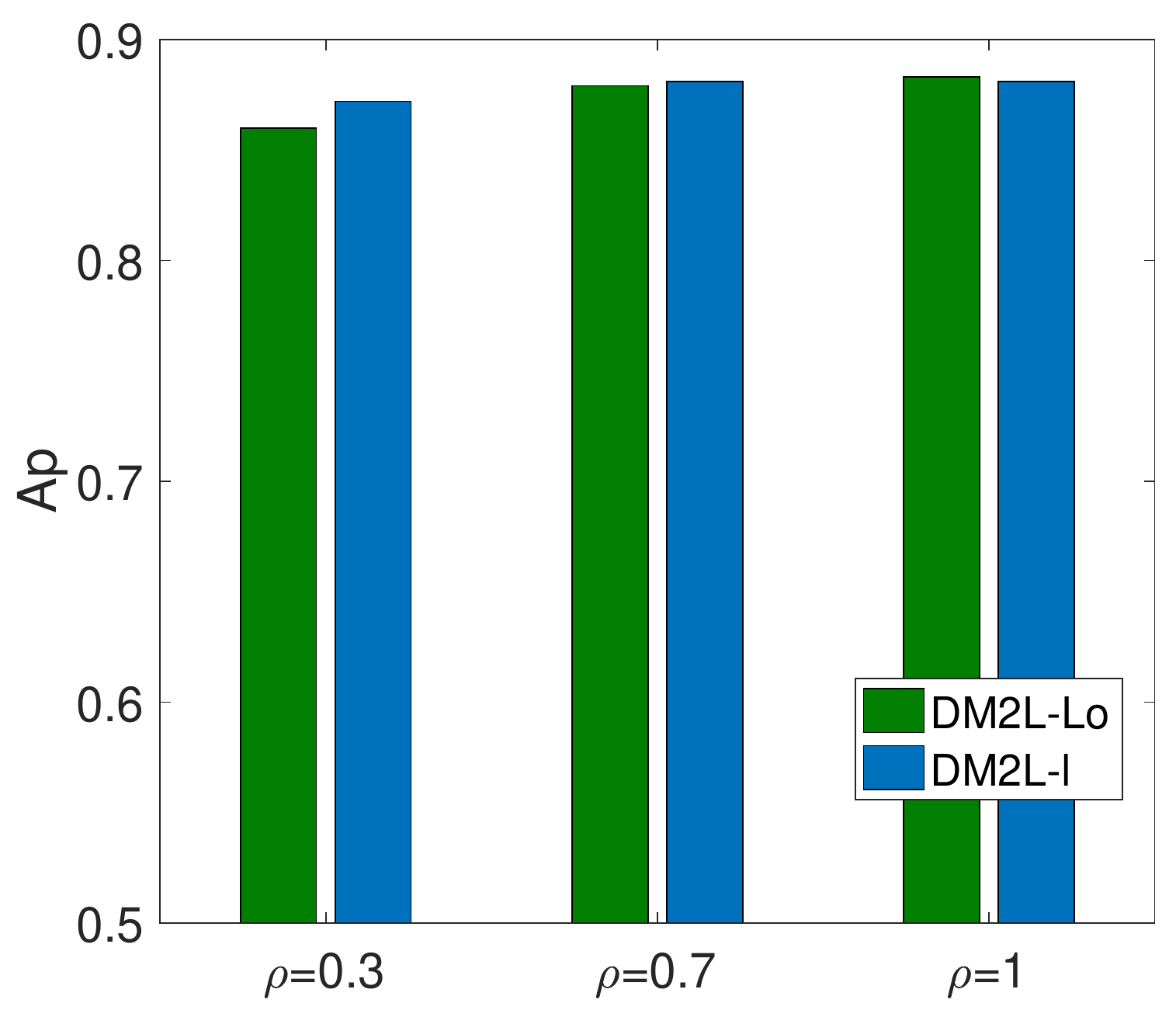}
	}%
	\centering
	\caption{DM2L-l vs. DM2L-Lo on Business dataset.}
	\label{disResu2}
\end{figure}

To show the effectiveness of discriminant label information, we compare DM2L-l to the above model DM2L-Lo on the Arts and Business datasets, and show the results in Fig. \ref{disResu1} and Fig. \ref{disResu2}. As can be seen, DM2L-l usually performs worse than DM2L-Lo on the full-label case, while DM2L-l usually performs better than DM2L-Lo on the missing-label case. We can see a similar phenomenon on the other datasets. This phenomenon indicates that when the label matrix is incomplete, the discriminant term can effectively avoid model over-fitting the local label structure. At the same time, this again reflects the importance of local label structure for multi-label classification, especially when the label matrix is complete.

\subsubsection{Convergence}

\begin{figure}[htbp!]
	\subfigure[Arts.]{
		\centering
		\includegraphics[width=4.5cm]{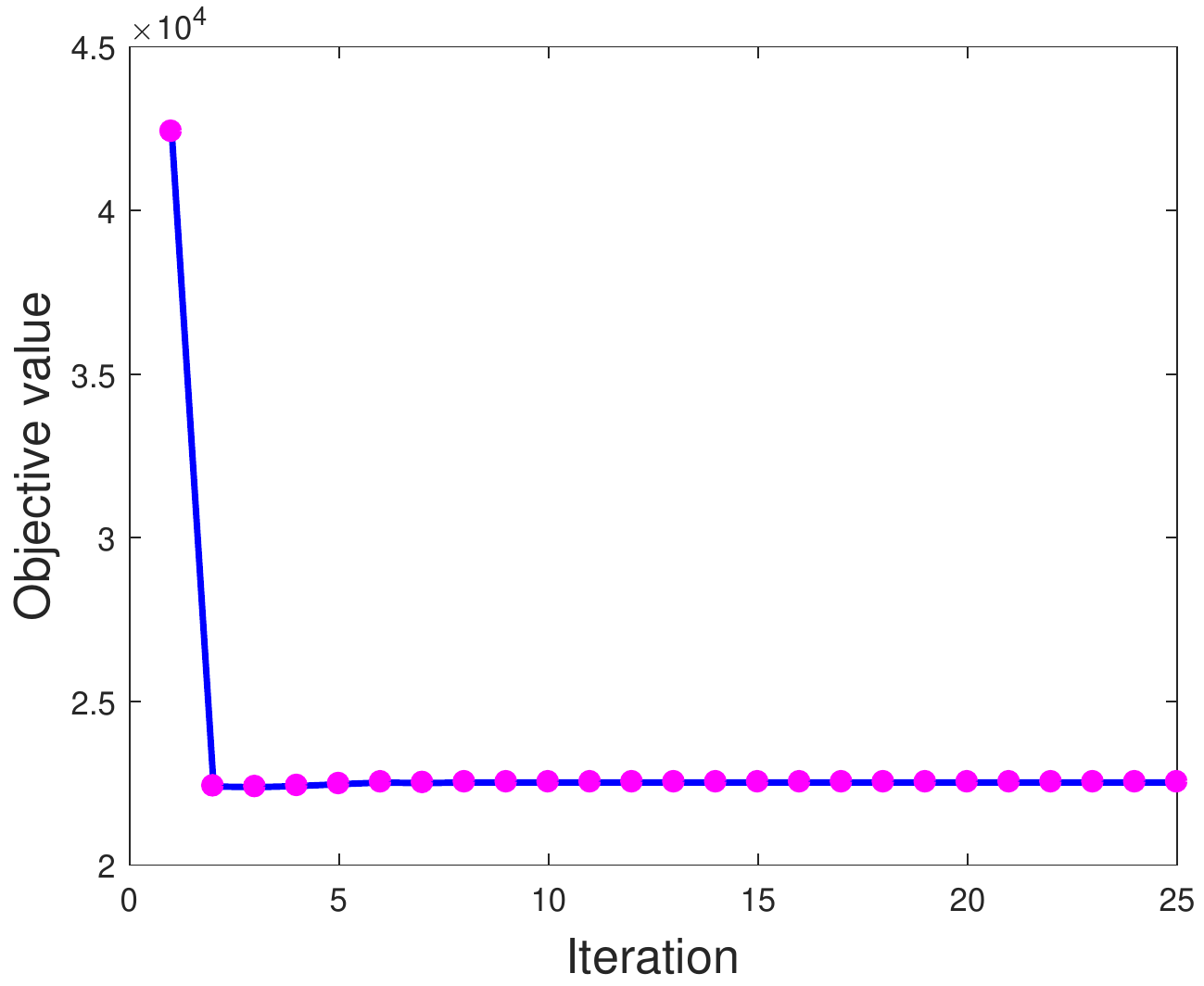}
	}%
	\subfigure[Business.]{
		\includegraphics[width=4.5cm]{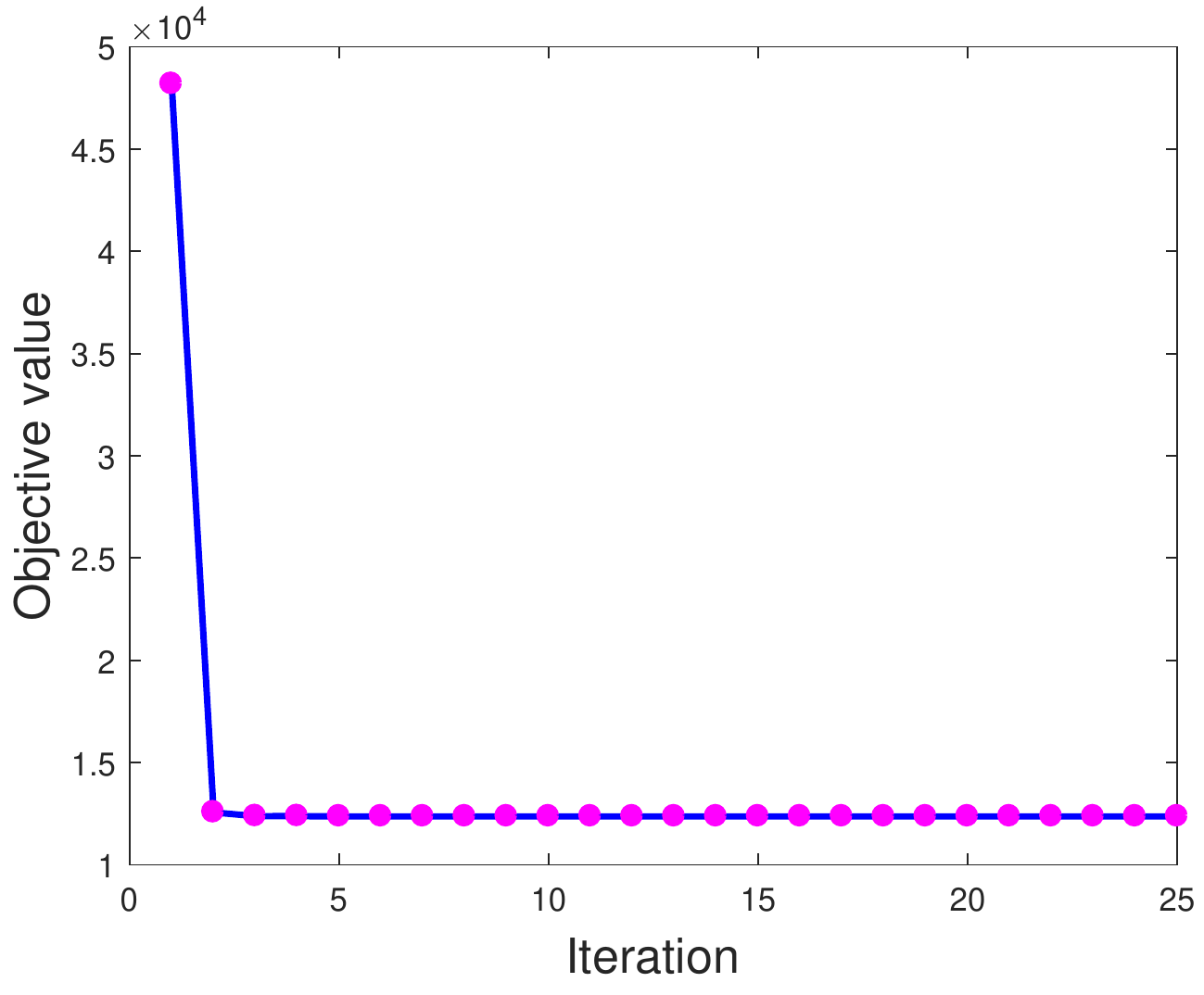}
	}%
	\centering
	\caption{Convergence of DM2L on the Arts and Business
		datasets.}
	\label{convergence_ana}
\end{figure}

We empirically study the convergence of DM2L, Fig. \ref{convergence_ana} shows the objective value w.r.t. the number of iterations for the full-label case on the Arts and Business. As can be seen, the objective converges quickly in a few iterations. A similar phenomenon can be observed on the other datasets.

\subsubsection{Hyper-parameter sensitivity analysis}
\begin{figure}[htbp!]
	\subfigure[]{
		\centering
		\includegraphics[height=3.7cm,width=4.5cm]{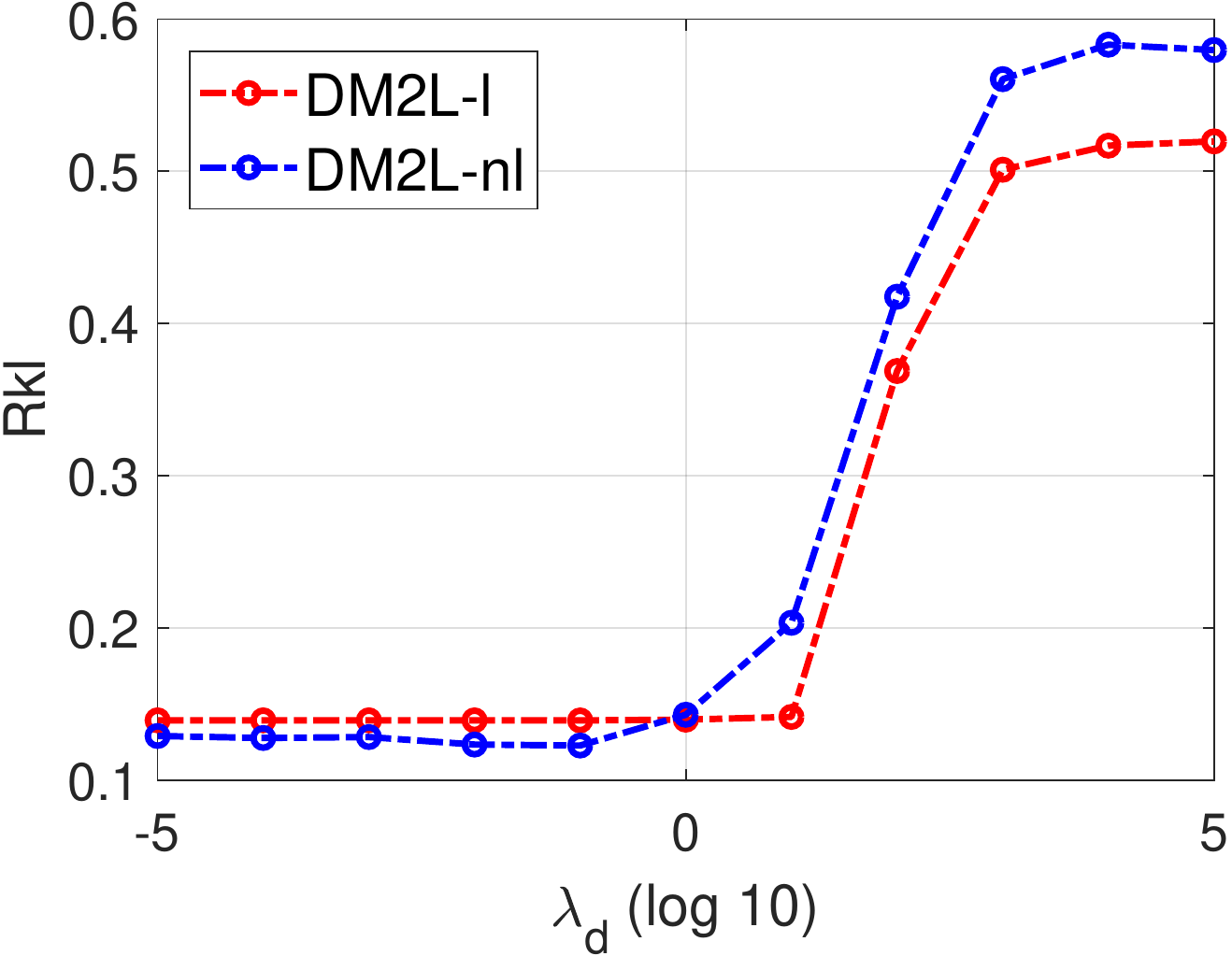}
	}%
	\subfigure[]{
		\includegraphics[height=3.7cm,width=4.5cm]{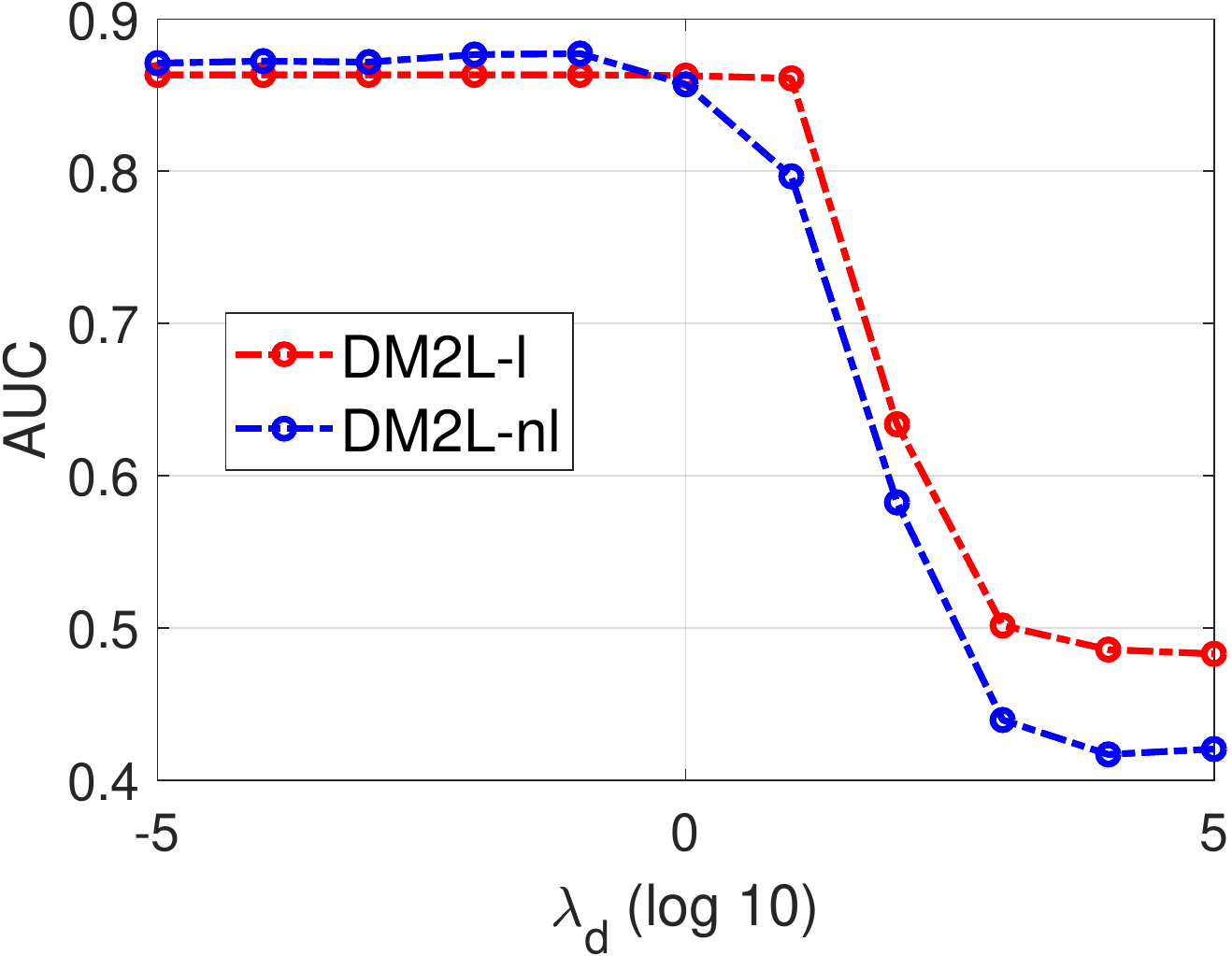}
	}%
	
	\subfigure[]{
		\includegraphics[height=3.7cm, width=4.5cm]{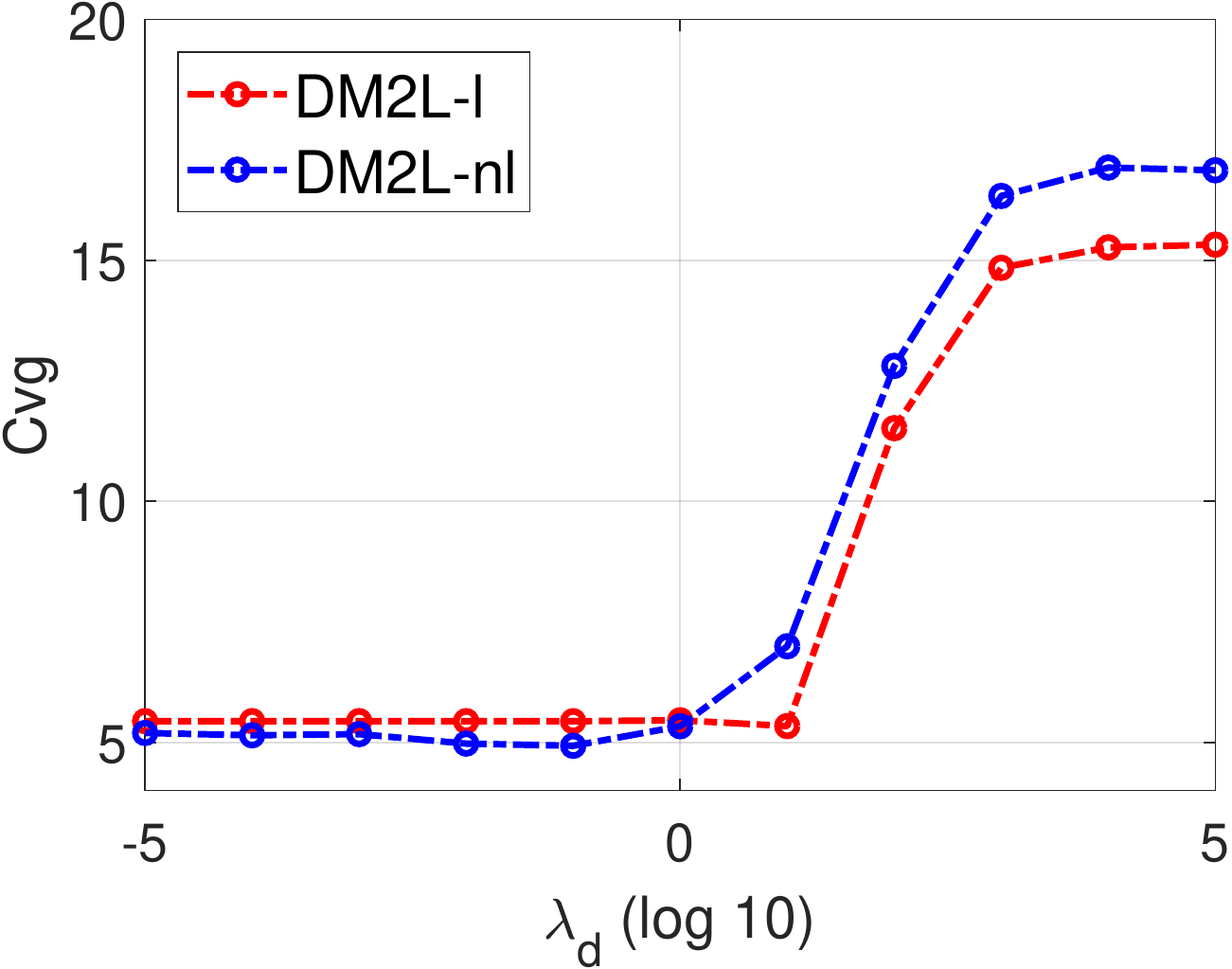}
	}%
	\subfigure[]{
		\includegraphics[height=3.7cm,width=4.5cm]{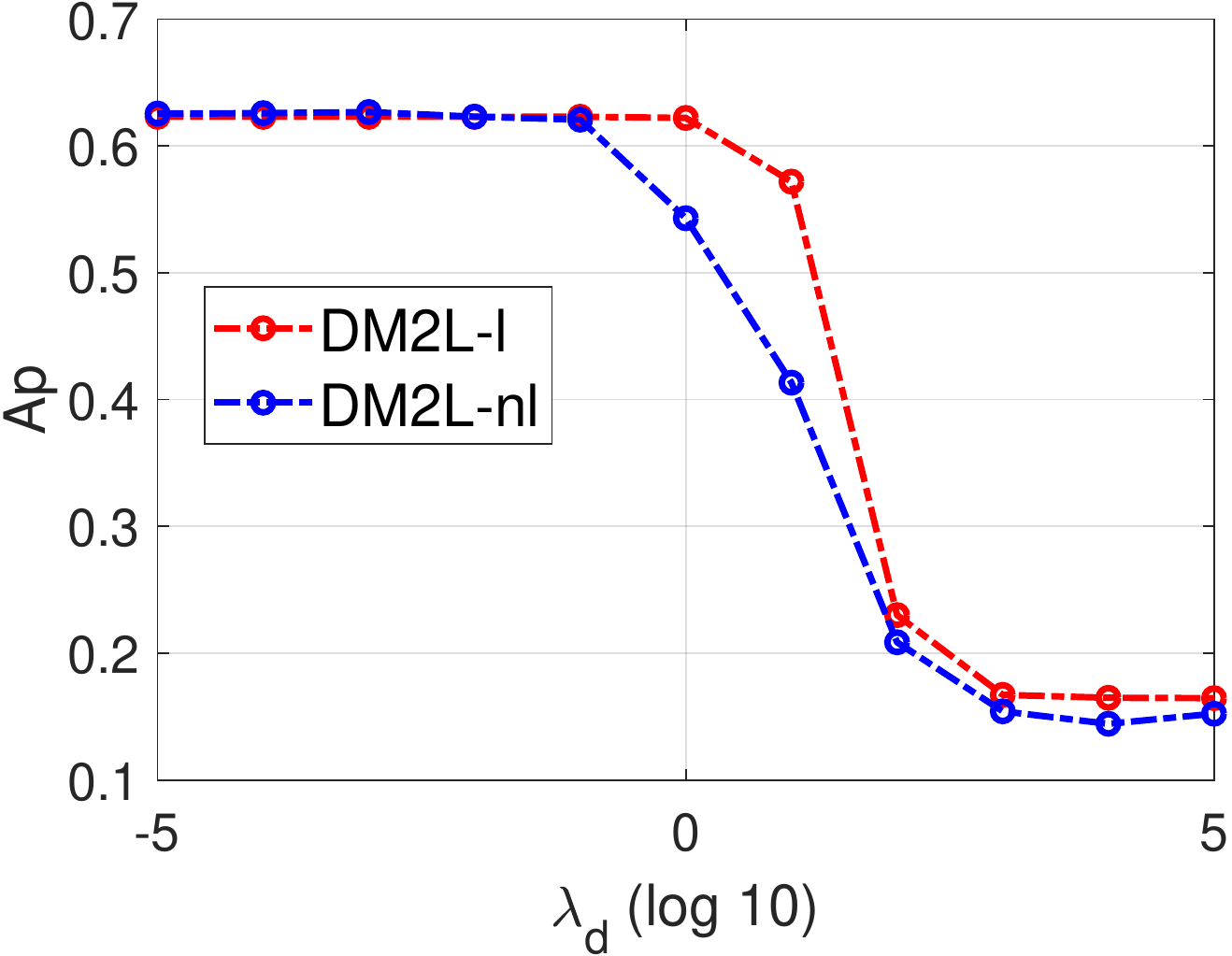}
	}%
	\centering
	\caption{Sensitivity analysis about $\lambda_d$ }
		\label{sen_r}
\end{figure}

\begin{figure}[htbp!]
	\subfigure[]{
		\centering
		\includegraphics[height=3.7cm,width=4.5cm]{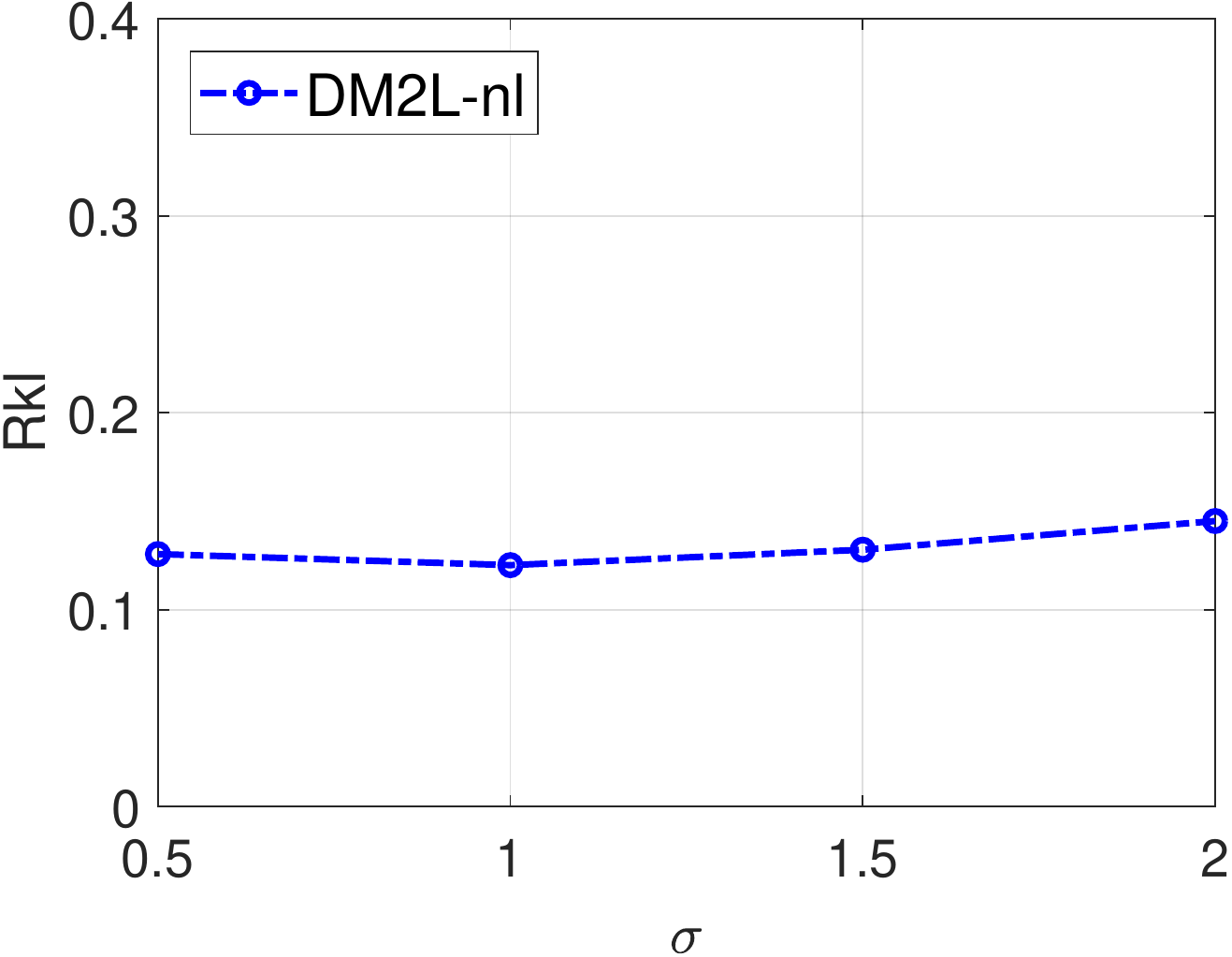}
	}%
	\subfigure[]{
		\includegraphics[height=3.7cm,width=4.5cm]{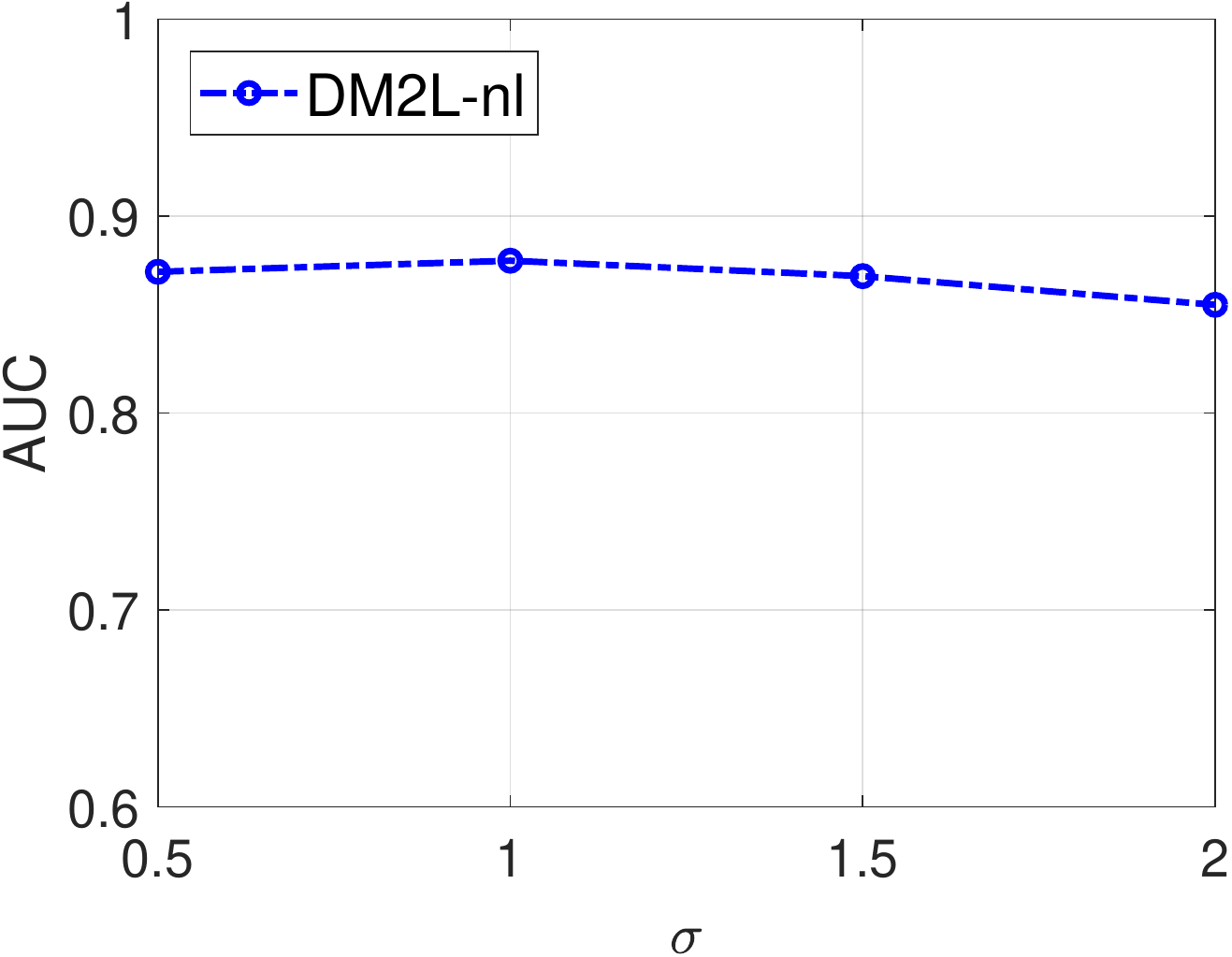}
	}%
	
	\subfigure[]{
		\includegraphics[height=3.7cm, width=4.5cm]{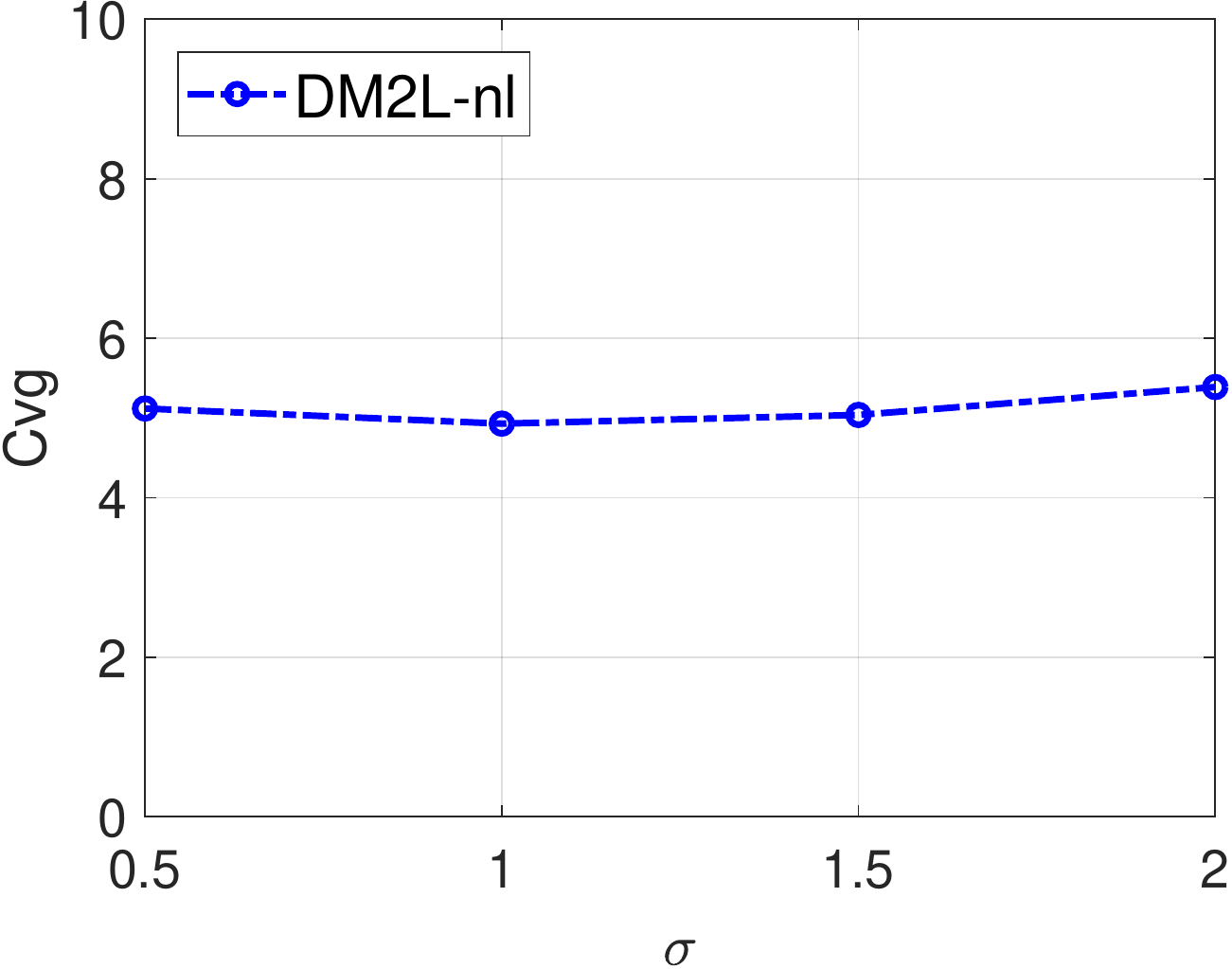}
	}%
	\subfigure[]{
		\includegraphics[height=3.7cm,width=4.5cm]{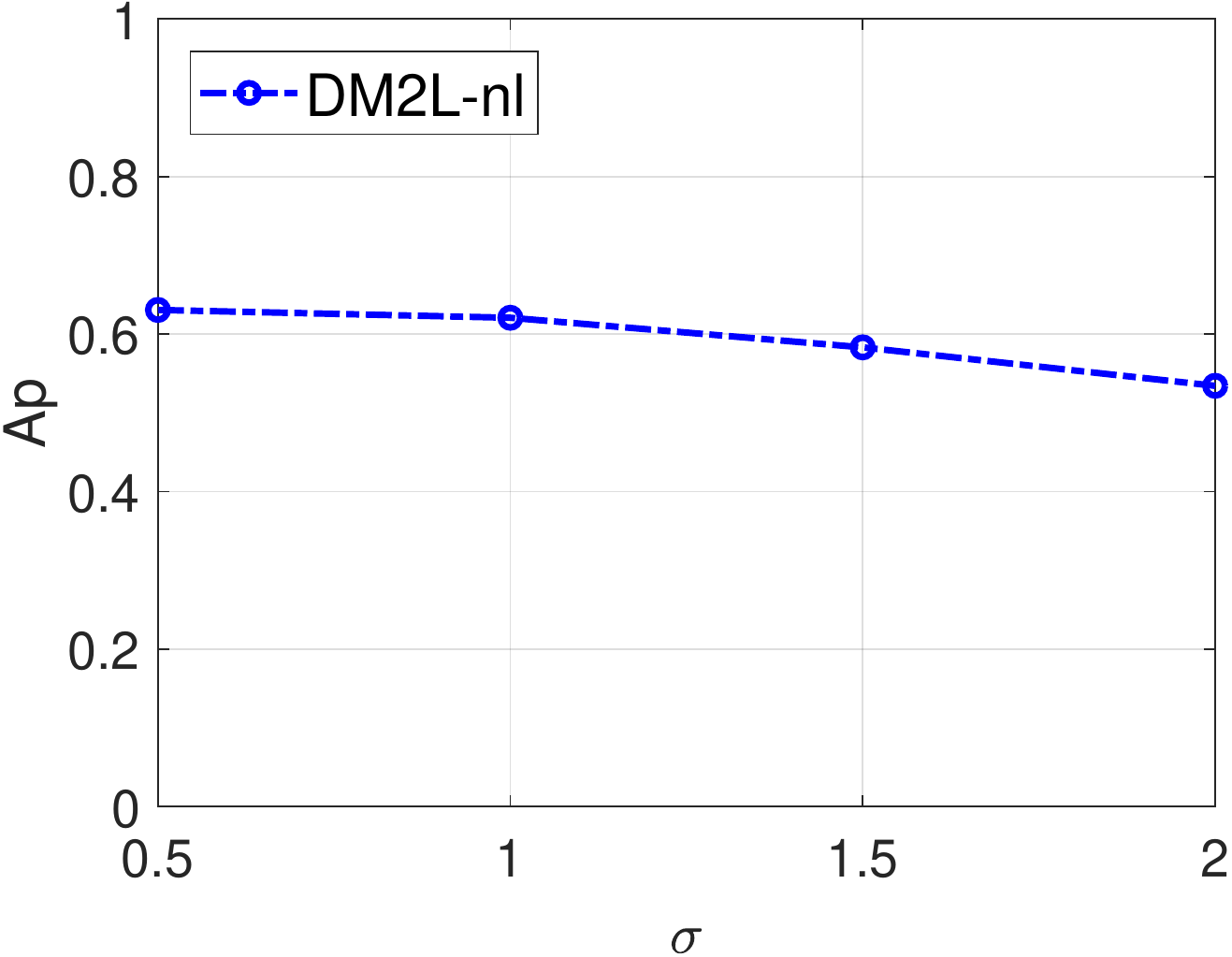}
	}%
	\centering
	\caption{Sensitivity analysis about $\delta$ }
		\label{sen_w}
\end{figure}

{{There is only one parameter, i.e., $\lambda_d$, in our proposed methods, which plays a role in trading off the loss function and the label structure regularizer. To analyze its sensitivity, we conduct experiments on the Arts dataset. The experimental results of D2ML-l and D2ML-nl with different values of $\lambda_d$ are depicted in Fig \ref{sen_r}.
From the experimental results, we note that both the performances of DM2L-l and DM2L-nl are relatively insensitive to the value of the regularization parameter $\lambda_d$. Additionally, we also analyze the sensitivity of implicit hyper-paramger of DM2L-nl, i.e., the Gaussian kernel width $\sigma$ in Eq.(\ref{gaussianF}), with fixing the value of $\lambda_d$ at the optimal value. The experimental results of D2ML-nl with different values of $\delta$ on the Arts dataset are depicted in Fig \ref{sen_w}.
From the experimental results, we note that the performance of DM2L-nl is also relatively insensitive to the values of Gaussian kernel width $\delta$. }}

\subsubsection{Time-comsuming comparison of different methods }
\begin{figure}[htbp!]
	\setlength{\abovecaptionskip}{0.2cm}
	\setlength{\belowcaptionskip}{0.cm}
	\centering
	\includegraphics[width=0.6\textwidth]{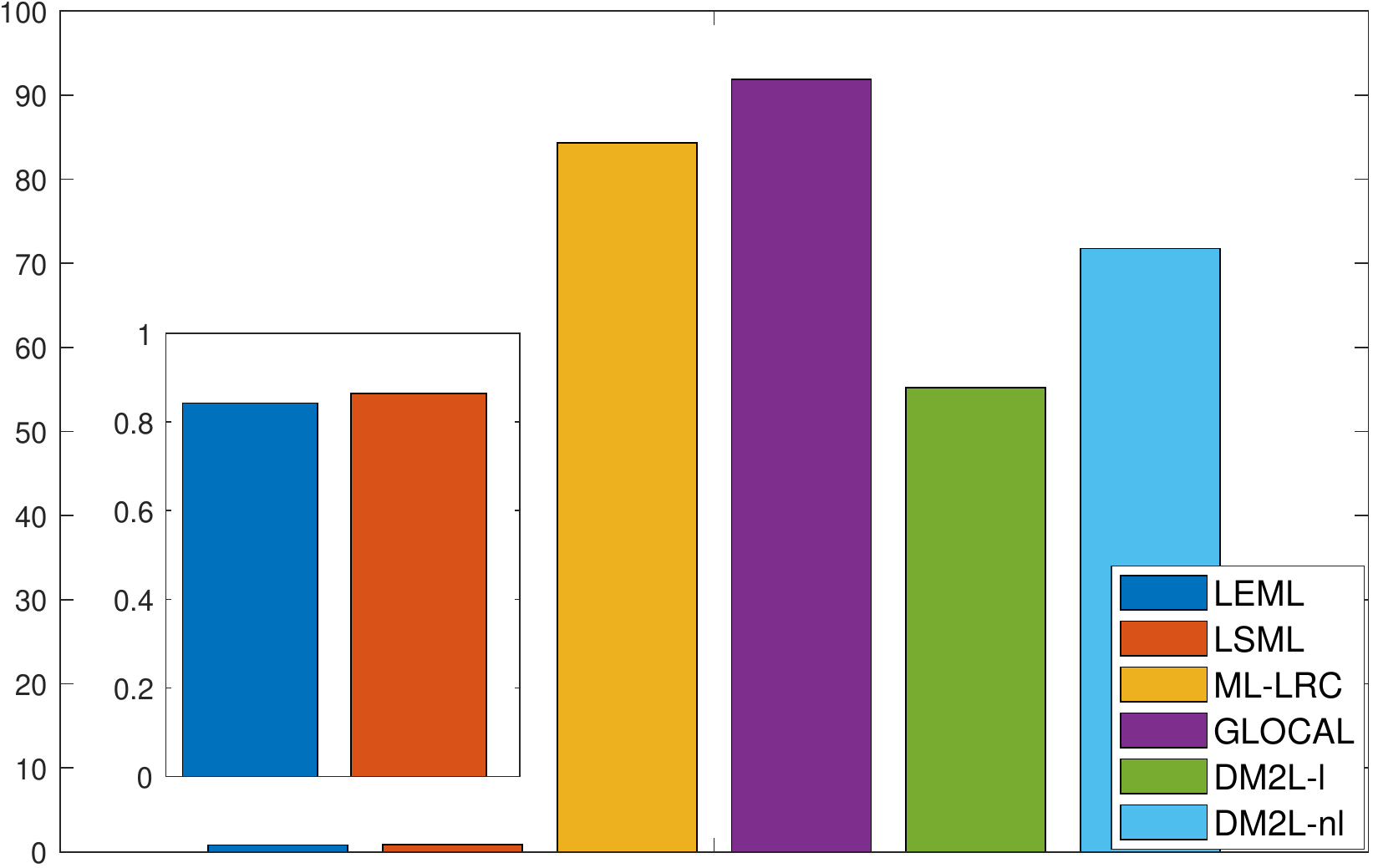}
	\caption{Running time results of different methods on the Arts dataset.}
	\label{timeofmethods}
\end{figure}

{{Besides the theoretical time complexity analysis in Section \ref{timeCom}, we also compare the time consumption of different methods in learning on the Arts dataset. From Fig. \ref{timeofmethods}, we can see that D2ML-l and D2ML-nl perform better than ML-LRC and GLOCAL. But they perform worse than LEML and LSML, which can be attributed to the complex CCCP optimization algorithm and will be addressed in the future research.}}

\section{Conclusion}
We proposed a simple yet effective discriminant multi-label learning method, i.e., DM2L, for multi-label learning with missing labels. Our method can jointly capture the global label structure, the local label structure and the discriminant information in a simple form. Additionally, we also provide a nonlinear extension via kernel trick to enhance DM2L and establish a concave-convex objective to learn these models. Compared with the previous work, our method is the first effective technique to explore local low-rank label structure for multi-label learning with missing labels and no longer suffers from the limitations of previous methods when exploring local label structure. It is one of the low-rank methods involving the fewest assumptions for multi-label learning with missing labels. Even so, extensive experiments show that our method still outperforms the state-of-the-art methods on learning with both full labels and missing labels. 

{{\textbf{Future work.} There are several directions deserved to explore further. First, when the dataset is very big, it would be interesting to generalize our learning technique to the online algorithm\cite{bottou2010large}, or consider using Fourier-transform techniques \cite{vedaldi2012efficient} and some randomized algorithms \cite{saibaba2017randomized} for fast computation. Second, we need to study the benefits of exploring local low-rank label structure in extreme multi-label learning problems, where the number of labels are extremely large. In extreme multi-label learning, the global low-rank label structure is usually violated due to the presence of tail labels\cite{xu2016robust}.}}

\section*{Acknowledgments}
This work is supported by the National Natural Science Foundation of China (NSFC) under the Grant No. 61672281. It is completed in the College of Computer Science and Technology at Nanjing University of Aeronautics and Astronautics.

\section*{References}

\bibliography{bare_jrnl_compsoc}

\begin{thebibliography}{10}
\expandafter\ifx\csname url\endcsname\relax
  \def\url#1{\texttt{#1}}\fi
\expandafter\ifx\csname urlprefix\endcsname\relax\def\urlprefix{URL }\fi
\expandafter\ifx\csname href\endcsname\relax
  \def\href#1#2{#2} \def\path#1{#1}\fi

\bibitem{boutell2004learning}
M.~R. Boutell, J.~Luo, X.~Shen, C.~M. Brown, Learning multi-label scene
  classification, Pattern recognition 37~(9) (2004) 1757--1771.

\bibitem{turnbull2008semantic}
D.~Turnbull, L.~Barrington, D.~Torres, G.~Lanckriet, Semantic annotation and
  retrieval of music and sound effects, IEEE Transactions on Audio, Speech, and
  Language Processing 16~(2) (2008) 467--476.

\bibitem{ueda2003parametric}
N.~Ueda, K.~Saito, Parametric mixture models for multi-labeled text, in:
  Advances in neural information processing systems, 2003, pp. 737--744.

\bibitem{tsoumakas2012introduction}
G.~Tsoumakas, M.-L. Zhang, Z.-H. Zhou, Introduction to the special issue on
  learning from multi-label data, Machine Learning 88~(1-2) (2012) 1--4.

\bibitem{zhang2007ml}
M.-L. Zhang, Z.-H. Zhou, Ml-knn: A lazy learning approach to multi-label
  learning, Pattern recognition 40~(7) (2007) 2038--2048.

\bibitem{clare2001knowledge}
A.~Clare, R.~D. King, Knowledge discovery in multi-label phenotype data, in:
  European Conference on Principles of Data Mining and Knowledge Discovery,
  Springer, 2001, pp. 42--53.

\bibitem{elisseeff2002kernel}
A.~Elisseeff, J.~Weston, A kernel method for multi-labelled classification, in:
  Advances in neural information processing systems, 2002, pp. 681--687.

\bibitem{read2011classifier}
J.~Read, B.~Pfahringer, G.~Holmes, E.~Frank, Classifier chains for multi-label
  classification, Machine learning 85~(3) (2011) 333.

\bibitem{furnkranz2008multilabel}
J.~F{\"u}rnkranz, E.~H{\"u}llermeier, E.~L. Menc{\'\i}a, K.~Brinker, Multilabel
  classification via calibrated label ranking, Machine learning 73~(2) (2008)
  133--153.

\bibitem{tsoumakas2007random}
G.~Tsoumakas, I.~Vlahavas, Random k-labelsets: An ensemble method for
  multilabel classification, in: European conference on machine learning,
  Springer, 2007, pp. 406--417.

\bibitem{zhou2017brief}
Z.-H. Zhou, A brief introduction to weakly supervised learning, National
  Science Review 5~(1) (2017) 44--53.

\bibitem{huang2015multi}
S.-J. Huang, S.~Chen, Z.-H. Zhou, Multi-label active learning: Query type
  matters., in: IJCAI, 2015, pp. 946--952.

\bibitem{gao2016multi}
N.~Gao, S.-J. Huang, S.~Chen, Multi-label active learning by model guided
  distribution matching, Frontiers of Computer Science 10~(5) (2016) 845--855.

\bibitem{yang2016improving}
H.~Yang, J.~T. Zhou, J.~Cai, Improving multi-label learning with missing labels
  by structured semantic correlations, in: European conference on computer
  vision, Springer, 2016, pp. 835--851.

\bibitem{wu2014multi}
B.~Wu, Z.~Liu, S.~Wang, B.~Hu, Q.~Ji, Multi-label learning with missing labels,
  in: 2014 22nd International Conference on Pattern Recognition, IEEE, 2014,
  pp. 1964--1968.

\bibitem{wu2015ml}
B.~Wu, S.~Lyu, B.~Ghanem, Ml-mg: Multi-label learning with missing labels using
  a mixed graph, in: Proceedings of the IEEE international conference on
  computer vision, 2015, pp. 4157--4165.

\bibitem{lin2013image}
Z.~Lin, G.~Ding, M.~Hu, J.~Wang, X.~Ye, Image tag completion via image-specific
  and tag-specific linear sparse reconstructions, in: Proceedings of the IEEE
  Conference on Computer Vision and Pattern Recognition, 2013, pp. 1618--1625.

\bibitem{chen2013fast}
M.~Chen, A.~Zheng, K.~Weinberger, Fast image tagging, in: International
  conference on machine learning, 2013, pp. 1274--1282.

\bibitem{cabral2011matrix}
R.~S. Cabral, F.~Torre, J.~P. Costeira, A.~Bernardino, Matrix completion for
  multi-label image classification, in: Advances in Neural Information
  Processing Systems, 2011, pp. 190--198.

\bibitem{goldberg2010transduction}
A.~Goldberg, B.~Recht, J.~Xu, R.~Nowak, J.~Zhu, Transduction with matrix
  completion: Three birds with one stone, in: Advances in neural information
  processing systems, 2010, pp. 757--765.

\bibitem{liu2015low}
M.~Liu, Y.~Luo, D.~Tao, C.~Xu, Y.~Wen, Low-rank multi-view learning in matrix
  completion for multi-label image classification, in: Twenty-Ninth AAAI
  Conference on Artificial Intelligence, 2015.

\bibitem{yuan2006model}
M.~Yuan, Y.~Lin, Model selection and estimation in regression with grouped
  variables, Journal of the Royal Statistical Society: Series B (Statistical
  Methodology) 68~(1) (2006) 49--67.

\bibitem{bucak2011multi}
S.~S. Bucak, R.~Jin, A.~K. Jain, Multi-label learning with incomplete class
  assignments, in: 2011 IEEE Conference on Computer Vision and Pattern
  Recognition (CVPR), IEEE, 2011, pp. 2801--2808.

\bibitem{zhu2018multi}
P.~Zhu, Q.~Xu, Q.~Hu, C.~Zhang, H.~Zhao, Multi-label feature selection with
  missing labels, Pattern Recognition 74 (2018) 488--502.

\bibitem{tan2017semi}
Q.~Tan, Y.~Yu, G.~Yu, J.~Wang, Semi-supervised multi-label classification using
  incomplete label information, Neurocomputing 260 (2017) 192--202.

\bibitem{liu2018svm}
Y.~Liu, K.~Wen, Q.~Gao, X.~Gao, F.~Nie, Svm based multi-label learning with
  missing labels for image annotation, Pattern Recognition 78 (2018) 307--317.

\bibitem{huang2019improving}
J.~Huang, F.~Qin, X.~Zheng, Z.~Cheng, Z.~Yuan, W.~Zhang, Q.~Huang, Improving
  multi-label classification with missing labels by learning label-specific
  features, Information Sciences 492 (2019) 124--146.

\bibitem{yu2014large}
H.-F. Yu, P.~Jain, P.~Kar, I.~Dhillon, Large-scale multi-label learning with
  missing labels, in: International conference on machine learning, 2014, pp.
  593--601.

\bibitem{xu2016robust}
C.~Xu, D.~Tao, C.~Xu, Robust extreme multi-label learning, in: Proceedings of
  the 22nd ACM SIGKDD international conference on knowledge discovery and data
  mining, 2016, pp. 1275--1284.

\bibitem{xu2014learning}
L.~Xu, Z.~Wang, Z.~Shen, Y.~Wang, E.~Chen, Learning low-rank label correlations
  for multi-label classification with missing labels, in: 2014 IEEE
  International Conference on Data Mining (ICDM), IEEE, 2014, pp. 1067--1072.

\bibitem{yu2018feature}
G.~Yu, X.~Chen, C.~Domeniconi, J.~Wang, Z.~Li, Z.~Zhang, X.~Wu, Feature-induced
  partial multi-label learning, in: 2018 IEEE International Conference on Data
  Mining (ICDM), IEEE, 2018, pp. 1398--1403.

\bibitem{tan2018incomplete}
Q.~Tan, G.~Yu, C.~Domeniconi, J.~Wang, Z.~Zhang, Incomplete multi-view
  weak-label learning., in: IJCAI, 2018, pp. 2703--2709.

\bibitem{strang1993introduction}
G.~Strang, G.~Strang, G.~Strang, G.~Strang, Introduction to linear algebra,
  Vol.~3, Wellesley-Cambridge Press Wellesley, MA, 1993.

\bibitem{lezama2018ole}
J.~Lezama, Q.~Qiu, P.~Mus{\'e}, G.~Sapiro, Ole: Orthogonal low-rank embedding,
  a plug and play geometric loss for deep learning, in: The IEEE Conference on
  Computer Vision and Pattern Recognition (CVPR), 2018.

\bibitem{huang2012multi}
S.-J. Huang, Z.-H. Zhou, Multi-label learning by exploiting label correlations
  locally, in: Twenty-sixth AAAI conference on artificial intelligence, 2012.

\bibitem{qiu2015learning}
Q.~Qiu, G.~Sapiro, Learning transformations for clustering and classification,
  The Journal of Machine Learning Research 16~(1) (2015) 187--225.

\bibitem{watson1992characterization}
G.~A. Watson, Characterization of the subdifferential of some matrix norms,
  Linear algebra and its applications 170 (1992) 33--45.

\bibitem{bottou2010large}
L.~Bottou, Large-scale machine learning with stochastic gradient descent, in:
  Proceedings of COMPSTAT'2010, Springer, 2010, pp. 177--186.

\bibitem{vedaldi2012efficient}
A.~Vedaldi, A.~Zisserman, Efficient additive kernels via explicit feature maps,
  IEEE transactions on pattern analysis and machine intelligence 34~(3) (2012)
  480--492.

\bibitem{saibaba2017randomized}
A.~K. Saibaba, A.~Alexanderian, I.~C. Ipsen, Randomized matrix-free trace and
  log-determinant estimators, Numerische Mathematik 137~(2) (2017) 353--395.

\end{thebibliography}

\end{document}